\documentclass[11pt]{article}

\usepackage[preprint]{acl}

\usepackage{microtype}
\usepackage{booktabs, array, makecell}
\usepackage{tabularray}
\UseTblrLibrary{booktabs}
\usepackage{amsmath}
\usepackage{amssymb}
\usepackage[ruled,vlined,linesnumbered]{algorithm2e}
\usepackage{dsfont}
\usepackage{nccmath}
\usepackage{graphicx}
\graphicspath{{figs/}}
 \usepackage{multirow}
 \usepackage{subcaption}
 \usepackage[most]{tcolorbox}

\usepackage{iftex}
\ifPDFTeX
  \usepackage[T1]{fontenc}
  \usepackage[utf8]{inputenc}
  \usepackage[english]{babel}
  \usepackage{tgtermes}
  \usepackage{tgcursor}
\else
\usepackage[english,bidi=default]{babel}
\babelfont{rm}[
  Extension=.otf,
  UprightFont=*-regular,
  BoldFont=*-bold,
  ItalicFont=*-italic,
  BoldItalicFont=*-bolditalic
]{texgyretermes}
\babelfont{tt}[
  Extension=.otf,
  UprightFont=*-regular,
  BoldFont=*-bold,
  ItalicFont=*-italic,
  BoldItalicFont=*-bolditalic
]{texgyrecursor}

\fi

\usepackage{tikz}
\definecolor{lightgreen}{RGB}{220,245,220}
\definecolor{darkgreen}{RGB}{0,100,0}

\usepackage{pifont}
\usepackage{xspace}
\usepackage{xcolor}
\usepackage{soul}
\usepackage[normalem]{ulem}
\usepackage{enumitem}

\newcolumntype{P}[1]{>{\centering\arraybackslash}p{#1}}
\newcolumntype{L}[1]{>{\raggedright\arraybackslash}p{#1}}

\usepackage{plume-style}

\newcommand{\task}{\rpsModule{MUSE}\xspace}
\newcommand{\benchmark}{{\mdseries\rpsBenchmark{MUSE-Bench}}\xspace}
\newcommand{\method}{\rpsMethod{PLUME}\xspace}

\newcommand{\gur}{\rpsModule{Global Update LoRA}\xspace}
\newcommand{\qame}{\rpsModule{Query-Activated Memory Evidence}\xspace}
\newcommand{\amd}{\rpsModule{Adaptive Memory Evidence Decoding}\xspace}
\newcommand{\llmname}[1]{\rpsModel{#1}\xspace}

\usepackage{colortbl} %
\definecolor{modelshade}{RGB}{242,246,250}
\definecolor{gaincolor}{RGB}{24,128,88}
\definecolor{vanillagray}{RGB}{90,90,90}

\definecolor{modelshade}{RGB}{238,244,250}

\newlength{\modelrowwidth}
\title{Towards Evolving Context Parameterization for Large Language Models}

\author{
 \textbf{Xiaobing Shi\textsuperscript{1,*}},
 \textbf{Zherui Li\textsuperscript{1,*}},
 \textbf{Yiming Jiang\textsuperscript{1}},
 \textbf{Kun Wang\textsuperscript{1,$\dagger$}},
 \textbf{Yufei Guo\textsuperscript{2,$\dagger$}}
\\
 \textsuperscript{1}Nanyang Technological University,
 \\
 \textsuperscript{2}Peking University\\
}

\makeatletter
\renewcommand\section{\@startsection{section}{1}{\z@}{-7pt plus -1pt minus -1pt}{4pt plus .5pt minus .5pt}{\large\bfseries\raggedright}}
\renewcommand\subsection{\@startsection{subsection}{2}{\z@}{-6pt plus -1pt minus -1pt}{3pt plus .5pt minus .5pt}{\normalsize\bfseries\raggedright}}
\renewcommand\paragraph{\@startsection{paragraph}{4}{\z@}{5pt plus 1pt minus 1pt}{-1em}{\normalsize\bfseries}}
\makeatother
\newcommand{\paperdisplayspacing}{%
  \setlength{\abovedisplayskip}{6pt plus 1pt minus 1pt}%
  \setlength{\belowdisplayskip}{6pt plus 1pt minus 1pt}%
  \setlength{\abovedisplayshortskip}{6pt plus 1pt minus 1pt}%
  \setlength{\belowdisplayshortskip}{6pt plus 1pt minus 1pt}}
\AddToHook{cmd/normalsize/after}{\paperdisplayspacing}

\begin{document}

\maketitle
\flushbottom
\paperdisplayspacing

\begin{abstract}

Context parameterization enables large language models (LLMs) to internalize contexts into reusable model parameters, avoiding repeated processing across subsequent queries.
However, existing methods typically assume static contexts and lack explicit mechanisms for distinguishing validity states under continual updates.
To study this real-world scenario, we formalized the \underline{M}emory \underline{U}pdating with \underline{S}equential \underline{E}volution \textbf{(\task{})} task and constructed \benchmark{} to evaluate update incorporation and unaffected-information preservation.
The resulting challenge requires preserving the global state while adjusting the contribution of memory evidence.
Motivated by this, we proposed \method{}, a training-free method that constructs a global update representation, activates memory evidence to form a local parameter view, and adaptively integrates their predictions during decoding.
Comprehensive evaluation on \benchmark{} demonstrated \method{}'s effectiveness in sequential evolution settings, yielding relative improvements of \textbf{29.9\%} in average ROUGE-L Recall and \textbf{54.9\%} in LLM-as-a-Judge.
Our codes are available at: \url{https://github.com/xiaobingshi-LLM/PLUME}.

\end{abstract}

\section{Introduction}
\label{sec:introduction}

Large Language Models (LLMs) demonstrated remarkable general-purpose capabilities in question answering, reasoning, and text generation, advancing natural language processing~\citep{brown2020language,wei2022chain,guo2025deepseek}. 
Many practical tasks are context-intensive, requiring models to efficiently locate and reuse information from long contexts across successive queries~\citep{bai2024longbench,bai2025longbench,li2025scbench}.
Existing retrieval-augmented generation (RAG)~\citep{lewis2020retrieval,guu2020retrieval} and long-context methods~\citep{ding2023longnet,chen2023extending} incur substantial inference overhead from repeated context processing~\citep{gao2023retrieval}, whereas context parameterization decouples this process from subsequent queries, enabling efficient reuse~\citep{snell2022learning,charakorn2026doc}.
Prior research followed two lines: \ding{172} one updated parametric knowledge through \textit{localized parameter modifications} without retraining the model, as in knowledge editing~\citep{mitchell2022fast,meng2022locating,meng2023massediting}; \ding{173} the other internalized contextual information into \textit{reusable parameter representations}, as in context distillation, Doc-to-LoRA, which compiled contexts into lightweight representations for subsequent queries~\citep{snell2022learning,charakorn2026doc}.

\begin{figure}[!t]
    \centering
    \includegraphics[
        width=\columnwidth,
        clip
    ]{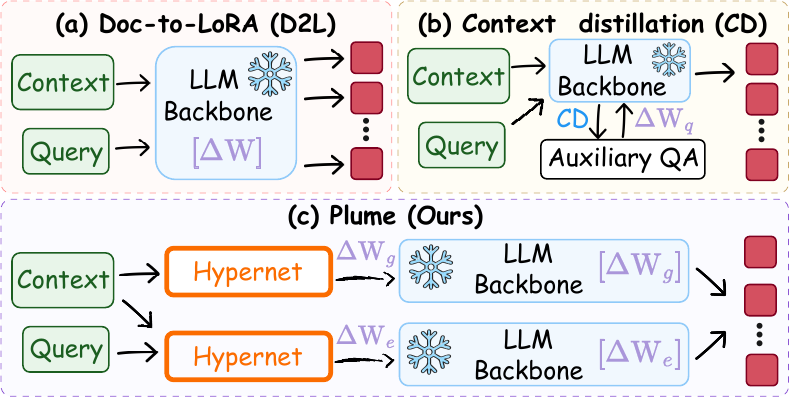}
    \caption{\textbf{Illustration of context parameterization under continual updates.} While existing methods perform well in conventional settings, they may conflate obsolete information with subsequent updates, whereas \method{} follows the currently valid information.
    }
    \label{fig:intro_motivation}
\end{figure}

Although these methods enabled effective, efficient context reuse, none addressed \textbf{parameterization under evolving information validity}~\citep{liu2025model,tian2026anyedit}.
In long-running agents and interactive systems, information continuously accumulates into persistent memory, making full context reconstruction impractical~\citep{park2023generative,packer2023memgpt,zhong2024memorybank}, motivating parameterized states that incorporate new information while preserving prior~\citep{jang2022temporalwiki}.
However, under continual updates, some information remains valid while other information is supplemented, replaced, or invalidated~\citep{wallat2026facts}, and jointly retaining information with different validity states can cause reliance on invalid information or inconsistent responses~\citep{marjanovic2024dynamicqa,wallat2026facts}.
Incorporating continual updates into a parameterized state while adhering to valid information thus remains an open challenge, as illustrated in Figure~\ref{fig:intro_motivation}: after an update, the parameterized state must \rpsStrongTerm{incorporate updates while preserving what remains unaffected}.

To study this continual-update setting, we formalized \textbf{Memory Updating with Sequential Evolution (\task{})} and designed a data construction pipeline preserving the original question and context structures, updating only the target and coupled information, and appending the updated context to the original.
This construction produces a context history with explicit update order and validity changes that supports the updated answer while no longer supporting the previous one.
Using this pipeline, we constructed \benchmark{} from five datasets, covering QA and reasoning.
Unlike prior work on conflicts between external context and model priors~\citep{xie2024adaptive,xu2024knowledge,kortukov2024studying}, \benchmark{} studies continual updates where \textbf{valid and invalid information coexist within a parameterized state}.
Experiments showed that parameterizing the full context into a unified representation substantially degraded performance, indicating severe interference among information with different validity states.

Building on \benchmark{} evaluation, we examined whether interference reflected \textit{information loss or merely suppressed activation} by computing token-level ROUGE-L Recall under teacher forcing for top-1 and top-5 candidate tokens at each step.
Top-5 recall reached \textbf{71.79}\%, substantially exceeding top-1's \textbf{45.90}\%, indicating that \textbf{valid information was largely retained in the output distribution but often outranked by invalid candidates}.
This finding motivates \method{}'s design: since the signal remains present but underweighted, we amplify the global update representation.
However, suppression varies across queries, and uniform amplification does not consistently identify which candidates require reweighting, motivating memory evidence as a local signal for targeted, per-query correction.
Memory evidence exhibits reduced reliability under phrasing divergence and often captures only partial information; we therefore incorporate it as a complement to the global representation rather than as a standalone signal.
\method{} therefore adaptively integrates both during decoding---\textit{the global representation as a stable prior} and \textit{memory evidence as a targeted correction}.

Experiments spanned multiple model families, including \llmname{Qwen3} and \llmname{Gemma2}~\citep{yang2025qwen3,team2024gemma}, on \benchmark{}, comparing \method{} with knowledge editing~\citep{jiang2025anyedit}, context distillation~\citep{askell2021general,snell2022learning}, Doc-to-LoRA~\citep{charakorn2026doc}, and other representative baselines. 
Averaged across \benchmark{}, \method{} improved ROUGE-L Recall from \textbf{44.46} (D2L) to \textbf{57.76} (\textbf{29.91\%}$\uparrow$), and LLM-as-a-Judge from \textbf{34.49} to \textbf{53.41} (\textbf{54.86\%}$\uparrow$), while consistently outperforming all baselines across models and task settings.
These results demonstrate that \method{} effectively mitigates parametric interference among information with different validity states while improving adherence to currently valid information throughout sequential evolution without sacrificing efficient context reuse.

\section{Background}
\label{sec:background}

We study context parameterization under continual updates, internalizing information into a parameterized state reused across queries as $C$ grows monotonically.
Let $C$, $q$, and $y$ denote the context, query, and answer, and $p_\theta(\cdot\mid\cdot)$ the conditional distribution of a $\theta$-parameterized language model.
Conventional long-context and RAG methods \textit{supply context at inference time}---concatenating context into the input window or retrieving passages---via $p_\theta(y\mid C,q)$~\citep{lewis2020retrieval,guu2020retrieval,ding2023longnet}, incurring reprocessing costs scaling with $|C|$, compounding across queries~\citep{gao2023retrieval}, and suffering position bias, underutilizing evidence away from the input's boundaries~\citep{liu2024lost,hsieh2024found}.
These costs motivate context parameterization internalizing $C$ into parameters decoupled from per-query inference.

Context Distillation (CD) instead transfers contextual information from explicit input into model parameters.
Given $C$ and its query distribution $\mathcal{Q}_C$, CD uses a teacher conditioned on $C$ to optimize a context-specific parameter state $\theta_C$:
\begin{equation}
    \min_{\theta_C}
    \mathbb{E}_{q\sim\mathcal{Q}_C}\!\left[
    D_{\mathrm{KL}}\!\left(
    p_\theta(\cdot\mid C,q)\,\|\,p_{\theta_C}(\cdot\mid q)
    \right)\right].
\label{eq:context-distillation}
\end{equation}
The teacher has direct access to $C$, whereas the student is optimized to reproduce the teacher distribution using only the query.
In this way, information originally supplied through the input context is encoded into the resulting parameter state.
After distillation, the model can answer queries without explicitly receiving $C$ during inference, allowing the same parameterized context to be reused across multiple queries~\citep{askell2021general,snell2022learning}.
However, CD requires \textbf{context-specific distillation data and iterative optimization} for each new context, making repeated updates costly.

Doc-to-LoRA (D2L) amortizes this per-context optimization using a hypernetwork $H_\phi$ that maps a context directly to a LoRA weight delta~\citep{hu2022lora,charakorn2026doc}:
\begin{equation} 
    \Delta \mathrm{W}_C=H_\phi(C),\qquad
    \theta_C=\theta\oplus\Delta \mathrm{W}_C .
\end{equation}
Here, $\Delta \mathrm{W}_C = B_C A_C$ denotes the effective low-rank weight update induced by the generated LoRA factors $(A_C,B_C)$, and $\oplus$ denotes applying this update to the frozen base model.
Instead of optimizing a separate parameter state for each context, D2L learns a generalizable mapping from contextual information to parameter updates across a collection of training contexts.
Once trained, $H_\phi$ generates a reusable context-specific adapter in \textbf{a single forward pass}, thereby substantially reducing the parameterization cost for previously unseen contexts.
The generated adapter can be independently applied to or removed from the frozen base model, providing a lightweight parametric representation of the corresponding context.
However, D2L parameterizes contexts without explicitly modeling continual updates to existing parameterized states.

\begin{figure*}[t]
    \centering
    \includegraphics[
        width=\textwidth,
        trim=0 0 16 0, %
        clip
    ]{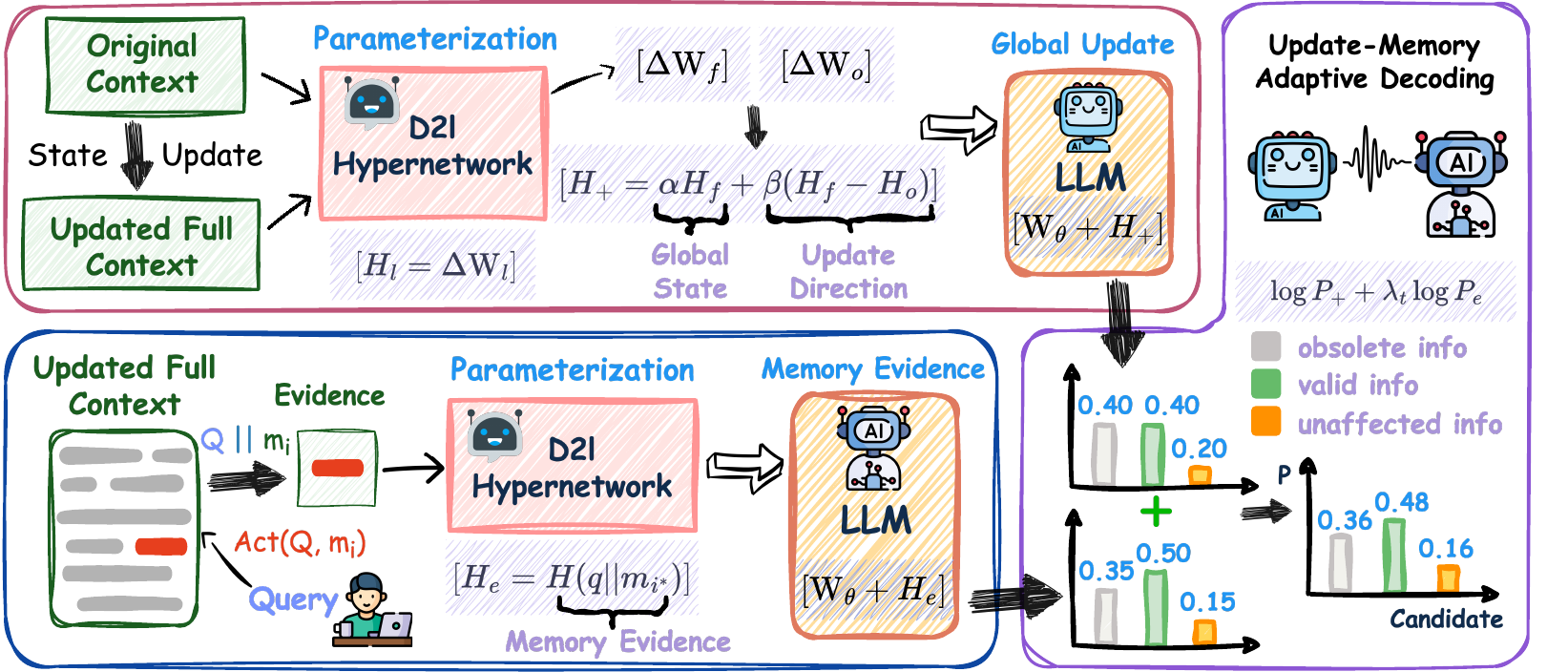}
    
    \caption{\textbf{The overall framework of our proposed \method{}.} \method{} constructs a global update representation from the full context state, activates memory evidence according to the query to provide a local parameter view, and adaptively integrates the global and local predictions during decoding.
    }
    \label{fig:method-overview}
\end{figure*}

\section{\benchmark{}}
\label{sec:benchmark}
To systematically evaluate context parameterization under continual updates, we defined the \textbf{\task{} (Memory Updating with Sequential Evolution)} task and constructed \benchmark{}, focusing on incorporating currently valid information while preserving unaffected information.

\subsection{Motivation and Task Formulation}
\paragraph{Motivation.}
Existing context parameterization methods primarily assume static, internally consistent contexts, without explicitly considering conflicts arising as contexts evolve~\citep{charakorn2026doc}.
Prior evaluations primarily assessed either static context internalization~\citep{snell2022learning,charakorn2026doc} or targeted factual modifications~\citep{meng2022locating,meng2023massediting}, leaving the incorporation of continual updates and preservation of unaffected information in parameterized representations insufficiently evaluated.
\benchmark{} is designed to address this gap in a unified setting.

\paragraph{Task Formulation.}
We formulate sequential evolution as a series of state transitions.
At update step $t$, let $C^{(t-1)}$ denote the context history preceding transition $t$, and let $C_u^{(t)}$ denote the update context.
The resulting context history is
\begin{equation}
    C^{(t)} = C^{(t-1)} \Vert C_u^{(t)},
    \label{eq:continual-update}
\end{equation}
where $C_u^{(t)}$ updates randomly sampled related information in $C^{(t-1)}$, while unaffected information remains valid.
The latest correction notice marks the state-transition boundary.
This formulation applies recursively: after step $t$, $C^{(t)}$ becomes the pre-update history for step $t+1$.
Let $\mathcal{A}$ map pre-update and current histories to a context-specific parameter state, optionally query-conditioned:
\begin{equation} 
    \theta^{(t)}(q)
    =
    \mathcal{A}\!\left(\theta, C^{(t-1)}, C^{(t)};q\right),
    \label{eq:context-parameterization}
\end{equation}
where $\theta$ denotes the base model parameters.
Context affects predictions only through the parameterized state $\theta^{(t)}(q)$. The target language model receives only $q$ as input, with no context text included in its input. 
The query may condition the construction or selection of the parameterized state.
\begin{equation} 
    \hat{y}^{(t)}(q)
    =
    \operatorname*{arg\,max}_{y} p_{\theta^{(t)}(q)}(y\mid q).
    \label{eq:muse-inference}
\end{equation}
Let $y^{\star(t)}(q)$ denote the reference answer supported by valid information in $C^{(t)}$.
At each transition, $\mathcal{Q}_{\mathrm{upd}}^{(t)}$ contains queries with update-affected answers, while $\mathcal{Q}_{\mathrm{keep}}^{(t)}$ contains queries with unchanged answers.
Both categories are evaluated under $C^{(t)}$ after parameterization.
\task{} evaluates whether the parameterized state incorporates the latest update while preserving unaffected information.

\subsection{Benchmark Construction}
\label{sec:benchmark-dataset}

To instantiate the transition-level formulation above, \benchmark{} designates one update transition per example and uses $C_o$, $C_u$, and $C_f$ as shorthand for $C^{(t-1)}$, $C_u^{(t)}$, and $C^{(t)}$, respectively.
It comprises five datasets: SQuAD, ROPES, 2WikiMultihopQA, MultiFieldQA-en, and QASPER~\citep{rajpurkar2016squad,lin2019reasoning,ho2020constructing,bai2024longbench,dasigi2021dataset}, with approximately \textbf{3,000 contexts} and \textbf{13,000 question-answer pairs}.
Given an original context $C_o={c_1,c_2,\ldots,c_n}$, we randomly sampled evidence units supporting a target answer and updated the corresponding facts while preserving the original question and context structures.
When other facts had semantic or reasoning dependencies on the target facts, we \textbf{updated them consistently} to construct an internally coherent update context $C_u$, then appended it to $C_o$ to form the full context history $C_f$.
The resulting context history supports the currently valid answers for update-affected queries while preserving the original answers for unaffected queries.

Candidate samples were generated with \llmname{GPT-series} models for fact modification, dependency identification, and question-answer annotation, primarily using \llmname{GPT-5.5}~\citep{openai2026gpt55} and \llmname{GPT-5.6-Sol} throughout the benchmark construction process.
We further used \llmname{GPT-5.6-Sol} to conduct 26 rounds of quality verification, with a subset of the annotations manually inspected in each round checking:
(1) whether $C_f$ supports the current correct answers;
(2) whether coupled information is consistently updated; and
(3) whether unaffected answers remain supported by $C_f$.
Detailed construction procedures, annotation prompts, and quality control protocols are provided in Appendix~\ref{app:annotation-guidelines}.

\subsection{Evaluation Protocol}
\label{sec:benchmark-evaluation}

\benchmark{} evaluates model outputs from four complementary perspectives of answer quality and update robustness.
\textbf{\rule[0.1em]{0.45em}{0.45em}\hspace{0.3em}D1 Answer Coverage}
uses token-level ROUGE-L Recall~\citep{lin2004rouge} to measure reference-answer coverage.
\textbf{\rule[0.1em]{0.45em}{0.45em}\hspace{0.3em}D2 Semantic Equivalence}
evaluates whether the generated answer is semantically equivalent to the reference answer.
\textbf{\rule[0.1em]{0.45em}{0.45em}\hspace{0.3em}D3 Final-Answer Agreement}
measures whether the final answer agrees with the reference answer.
\textbf{\rule[0.1em]{0.45em}{0.45em}\hspace{0.3em}D4 Locality}
measures whether the model preserves correct answers for queries unaffected by the update.
Together, these metrics evaluate both effective incorporation of valid information and reliable preservation of unaffected information.

All metrics are computed at the query level and then aggregated within each dataset for overall performance comparison.
For D2 and D3, we used \textbf{\llmname{GPT-5.6-Sol}} as the LLM Judge with unified evaluation prompts and consistent evaluation criteria.
Detailed evaluation settings and complete evaluation rubrics are provided in Appendix~\ref{app:evaluation-setup}.

\section{Method}
\label{sec:method}

We proposed \textbf{\underline{P}arametric \underline{L}oRA \underline{U}pdates with \underline{M}emory \underline{E}vidence (\method{})}, which combines a global update representation, query-activated memory evidence, and adaptive decoding for context parameterization under continual updates. 
Figure~\ref{fig:method-overview} illustrates the overall framework.

\subsection{Global Update Representation}
\label{sec:global_update_backbone}

Throughout sequential evolution, a parameterized context state can simultaneously entangle currently valid, invalidated, and unaffected information without an explicit mechanism for disentangling them.
Given the full context history $C_{\mathrm{full}}$ and its pre-update history $C_{\mathrm{old}}$, we parameterize both separately with the pretrained D2L hypernetwork $H_\phi$ introduced in Section~\ref{sec:background}, obtaining $\Delta \mathrm{W}_{\mathrm{old}}=H_\phi(C_{\mathrm{old}})$ and $\Delta \mathrm{W}_{\mathrm{full}}=H_\phi(C_{\mathrm{full}})$.
The global update representation is accordingly defined to jointly integrate these two parameterized states as
\begin{equation}
\Delta \mathrm{W}_g = \alpha \Delta \mathrm{W}_{\mathrm{full}} + \beta\left(\Delta \mathrm{W}_{\mathrm{full}}-\Delta \mathrm{W}_{\mathrm{old}}\right),
\label{eq:global_adapter}
\end{equation}
where $\alpha,\beta\geq0$ are weighting coefficients: the first term \textbf{retains the complete context state}, and the second \textbf{selectively amplifies the parameter shift induced by the latest update} without discarding it.
Operationally, we implement Eq.~\ref{eq:global_adapter} by concatenating the LoRA factors along the rank dimension, yielding a global adapter with rank $r_g=r_{\mathrm{full}}+r_{\mathrm{old}}$.
Appendix~\ref{app:lora-composition} provides the exact construction and its extension to chunked contexts.
The full-context component provides the broader context required for answering queries, including information unaffected by the update, while the difference term emphasizes changes between the pre-update and full parameterized states. Using the difference term alone would omit this explicit source of contextual support, since the parameter shift need not encode all background information required to interpret and answer a query.
We treat $\alpha$ and $\beta$ as tunable weighting coefficients that favor \textit{a supplementary rather than substitutive role} for the update-shift term, a design choice empirically validated in Section~\ref{sec:experiments}.
At generation step $t$, its predictive distribution is
\begin{equation}
p_{g,t}(v) = p_{\theta\oplus\Delta \mathrm{W}_g} \left(v\mid q,y_{<t}\right),
\label{eq:global-distribution}
\end{equation}
where $v$ denotes a candidate token, $y_{<t}$ is the generated prefix, and $\oplus$ denotes applying the corresponding parameter update to the frozen base model.

\subsection{Query-Activated Memory Evidence}
\label{sec:memory_evidence_activation}

While the global update representation aggregates the entire context history into a parameterized state, this compression can attenuate fine-grained detail in long, information-dense contexts.
We therefore construct \textbf{a complementary local parameter view} by segmenting $C_{\mathrm{full}}$ into memory units and scoring lexical relevance via query-term coverage and saturated term-frequency matching.
Let $s_{\max}$ denote the highest score among candidate units.
Among units whose scores fall within a margin $\delta$ of $s_{\max}$, we activate the temporally latest as local memory evidence $m^\star(q)$, where $\delta \geq 0$ controls how close in score two units must be for temporal order to take precedence, so that \textbf{recency breaks ties only among comparably relevant candidates} rather than overriding a clearly stronger lexical match.
Temporal order follows each unit's position within $C_{\mathrm{full}}$, with $C_{\mathrm{update}}$ units treated as later than $C_{\mathrm{old}}$ units.
The memory evidence is parameterized as
\begin{equation}
\Delta \mathrm{W}_e(q)=H_\phi\left(m^\star(q)\right),
\label{eq:evidence_adapter}
\end{equation}
with the corresponding predictive distribution
\begin{equation}
p_{e,t}(v) = p_{\theta\oplus\Delta \mathrm{W}_e(q)} \left(v\mid q,y_{<t}\right).
\label{eq:evidence-distribution}
\end{equation}
Formed independently of the compressed global representation, $\Delta \mathrm{W}_e(q)$ internalizes a localized memory state that surfaces under-represented detail, reinforcing rather than replacing it.
The query determines which evidence is selected, but \textbf{the hypernetwork receives only the selected evidence text}; the dependence of $\Delta \mathrm{W}_e(q)$ on $q$ therefore arises through evidence selection. During generation, this activated memory evidence influences predictions through the resulting parameter adapter.

\begingroup

\definecolor{headerbg}{HTML}{F1F8FE}

\definecolor{recallarrow}{HTML}{F05A3C}

\definecolor{llmarrow}{HTML}{3182CE}

\newcommand{\RecallMetric}{Recall\,\textcolor{recallarrow}{$\uparrow$}}

\newcommand{\LLMMetric}{LLM\,\textcolor{llmarrow}{$\uparrow$}}

\newcommand{\PLUMENumber}[1]{{\color{black}\bfseries #1}}

\newcommand{\SecondNumber}[1]{\uline{#1}}

\newcommand{\MethodFont}{\fontsize{12pt}{34pt}\selectfont\bfseries}

\newlength{\ModelHeadingOffset}

\newlength{\ModelHeadingHeight}

\newlength{\ModelHeadingAboveSep}

\newlength{\ModelHeadingBelowSep}

\setlength{\ModelHeadingOffset}{10.1cm}

\setlength{\ModelHeadingHeight}{8pt}

\setlength{\ModelHeadingAboveSep}{0pt}

\setlength{\ModelHeadingBelowSep}{0pt}

\newcommand{\ModelHeadingText}[1]{%
  \makebox[0pt][l]{\hspace*{\ModelHeadingOffset}%
    {\color{gray}\itshape #1}}%
}

\newcommand{\DatasetHeader}[1]{%
  \begin{tblr}{
      colspec={Q[c,wd=1.35cm]Q[c,wd=1.05cm]},
      colsep=3.1pt,
      rowsep=2pt,
      rows={font=\bfseries,halign=c,valign=m},
      hline{2}={wd=0.5pt,solid,leftpos=-0.5,rightpos=-0.5},
    }
    \SetCell[c=2]{halign=c,valign=m}#1 & {} \\
    \RecallMetric & \LLMMetric
  \end{tblr}%
}

\begin{table*}[t]

\centering

\small

\resizebox{\textwidth}{!}{%

\begin{tblr}{
    colspec = {
        Q[l,wd=4.6cm,valign=m]
        *{5}{
            Q[c,wd=1.35cm,valign=m]
            Q[c,wd=1.05cm,valign=m]
        }
    },
    rowsep = 1pt,
    rulesep = 2pt,
    colsep = 4pt,
    row{1} = {bg=headerbg, halign=c, valign=m, font=\bfseries,
              abovesep=2pt, belowsep=2pt},
    hline{1} = {wd=2pt, solid},
    hline{2,11} = {wd=2pt, solid},
    hline{10,19} = {1}{-}{wd=0.5pt, solid},
    hline{10,19} = {2}{-}{wd=0.5pt, solid},
    column{1} = {font=\bfseries},
    cell{3-5,12-14}{1-11} = {fg=gray},
}

\SetCell{halign=c,valign=m}{{\MethodFont Method}}

& \SetCell[c=2]{halign=c,valign=m}{\DatasetHeader{SQuAD}} & {}

& \SetCell[c=2]{halign=c,valign=m}{\DatasetHeader{ROPES}} & {}

& \SetCell[c=2]{halign=c,valign=m}{\DatasetHeader{2Wiki}} & {}

& \SetCell[c=2]{halign=c,valign=m}{\DatasetHeader{MFQA}} & {}

& \SetCell[c=2]{halign=c,valign=m}{\DatasetHeader{QASPER}} & {} \\

\SetRow{ht=\ModelHeadingHeight,
        abovesep=\ModelHeadingAboveSep,
        belowsep=\ModelHeadingBelowSep}
\SetCell[c=11]{halign=l,valign=m}{\ModelHeadingText{Qwen-3}} \\

\SetHline{1-11}{0.5pt,solid}

Base w/ Context (oracle)

& 95.86 & 91.69

& 97.82 & 95.91

& 60.11 & 52.35

& 64.50 & 60.08

& 61.75 & 53.95

\\

CD (oracle)

& 94.93 & 91.75

& 98.31 & 97.52

& 68.51 & 65.44

& 66.24 & 61.17

& 62.46 & 55.81

\\

D2L (oracle)

& 76.77 & 64.92

& 85.57 & 80.39

& 43.13 & 38.33

& 28.80 & 20.34

& 22.98 & 15.80

\\

\SetHline{1-11}{0.5pt,solid}

Base w/o Context

& 24.61 & 15.19

& 82.88 & 64.99

& 31.75 & 24.67

& 22.48 & 11.33

& 1.28 & 1.89

\\

AnyEdit

& 15.77 & 12.44

& 76.43 & 67.48

& 16.25 & 15.00

& 13.23 & 9.25

& 7.40 & 4.02

\\

CD

& 40.57 & 27.85

& 74.69 & 67.00

& 36.15 & 30.67

& \SecondNumber{30.78} & \SecondNumber{23.33}

& 16.78 & 9.57

\\

D2L

& \SecondNumber{54.96} & \SecondNumber{37.27}

& \SecondNumber{86.99} & \SecondNumber{81.33}

& \SecondNumber{36.30} & \SecondNumber{30.84}

& 27.15 & 12.00

& \SecondNumber{16.88} & \SecondNumber{11.02}

\\

\SetRow{font=\bfseries\color{black}}

\method{}

& \PLUMENumber{80.65} & \PLUMENumber{76.81}

& \PLUMENumber{94.61} & \PLUMENumber{90.26}

& \PLUMENumber{47.24} & \PLUMENumber{41.48}

& \PLUMENumber{37.76} & \PLUMENumber{34.01}

& \PLUMENumber{28.53} & \PLUMENumber{24.48}

\\

\SetRow{ht=\ModelHeadingHeight,
        abovesep=\ModelHeadingAboveSep,
        belowsep=\ModelHeadingBelowSep}
\SetCell[c=11]{halign=l,valign=m}{\ModelHeadingText{Gemma-2}} \\

\SetHline{1-11}{0.5pt,solid}

Base w/ Context (oracle)

& 90.73 & 84.65

& 84.73 & 80.17

& 36.39 & 28.21

& 31.81 & 27.49

& 52.48 & 44.72

\\

CD (oracle)

& 89.92 & 84.77

& 85.81 & 79.38

& 42.23 & 34.39

& 35.11 & 30.84

& 52.46 & 45.40

\\

D2L (oracle)

& 70.81 & 57.46

& 58.84 & 50.43

& 22.11 & 16.15

& 14.75 & 7.72

& 19.10 & 12.58

\\

\SetHline{1-11}{0.5pt,solid}

Base w/o Context

& 22.07 & 12.54

& 44.44 & 25.95

& 14.76 & 6.49

& 5.02 & 3.36

& 0.95 & 1.17

\\

AnyEdit

& 16.24 & 8.93

& 51.81 & 41.72

& 10.37 & 8.64

& 5.23 & 4.06

& 0.54 & 0.46

\\

CD

& 32.30 & 20.19

& 54.63 & 45.21

& 20.82 & 14.48

& 9.72 & \SecondNumber{6.76}

& 12.94 & 5.23

\\

D2L

& \SecondNumber{49.17} & \SecondNumber{31.36}

& \SecondNumber{57.53} & \SecondNumber{48.69}

& \SecondNumber{21.09} & \SecondNumber{15.96}

& \SecondNumber{11.44} & 6.55

& \SecondNumber{14.22} & \SecondNumber{7.96}

\\

\SetRow{font=\bfseries\color{black}}

\method{}

& \PLUMENumber{74.80} & \PLUMENumber{68.60}

& \PLUMENumber{62.70} & \PLUMENumber{55.78}

& \PLUMENumber{29.85} & \PLUMENumber{24.82}

& \PLUMENumber{23.29} & \PLUMENumber{19.42}

& \PLUMENumber{24.51} & \PLUMENumber{20.35}

\\

\SetHline{1-11}{2pt,solid}

\end{tblr}%

}

\caption{\textbf{A Comprehensive Comparison of Different Methods.} ROUGE-L Recall and LLM-as-a-Judge are reported on MUSE-Bench datasets, where \textbf{bold} and \underline{underline} denote the best and second-best values respectively.}

\label{tab:main_results}

\end{table*}

\endgroup

\subsection{Adaptive Memory Evidence Decoding}
\label{sec:adaptive_evidence_decoding}

Here, $p_{g,t}$ characterizes the global update state, while $p_{e,t}$ captures the query-activated memory evidence.
Since the relative reliability of each context can vary unpredictably across generation steps depending on its alignment with the query, we adaptively reweight the evidence term based on its token-level divergence from the global prediction, formally quantified via the JS divergence:
\begin{equation} 
d_t = D_{\mathrm{JS}}\left(p_{e,t}\parallel p_{g,t}\right),
\label{eq:js-divergence}
\end{equation}
A larger $d_t$ indicates that the activated memory evidence departs substantially from the global prediction at step $t$, suggesting complementary information worth emphasizing. 
We therefore map $d_t$ to an adaptive gating weight that increases monotonically with divergence and saturates at $\lambda_{\max}$:
\begin{equation}
\lambda_t = \lambda_{\max}\frac{d_t}{d_t+\tau},
\label{eq:adaptive_gate}
\end{equation}
where $\lambda_{\max}$ bounds the influence of the memory evidence, preventing the model from degenerating into a purely local decoder under sharp divergence, and $\tau>0$ controls the sensitivity of $\lambda_t$ to $d_t$.
Guided by this gate, we \textbf{retain the global prediction as decoding backbone} and fuse in this evidence through
\begin{equation}
S_t(v) = \log p_{g,t}(v) + \lambda_t\log p_{e,t}(v),
\label{eq:fusion-score}
\end{equation}
with the final token selected as $y_t=\arg\max_v S_t(v)$ at each generation step.
As $\lambda_t$ approaches zero, the fused scores approach the global log-probabilities, making token selection increasingly governed by the global prediction. Increasing $\lambda_t$ gives greater weight to the local evidence's relative token preferences, allowing it to more strongly reshape the ranking of candidate tokens.
Given its sensitivity to phrasing variation and partial coverage, memory evidence serves as a complementary signal to the global representation, preventing any single source from dominating the fused prediction.

\section{Experiments}
\label{sec:experiments}

We evaluated \method{} on \benchmark{}, cross-task generalization, efficiency and information retention, component ablations, and representative cases.

\subsection{Experimental Setup}
\label{sec:exp_setup}

\paragraph{Datasets and Models.}
We evaluated on the five \benchmark{} datasets and used GSM8K~\citep{cobbe2021training} and CRUXEval~\citep{gu2024cruxeval} for cross-task generalization, with \llmname{Qwen3} and \llmname{Gemma2} as model backbones.
Both D2L and \method{} used the official pretrained D2L hypernetwork checkpoint for context parameterization. 

\paragraph{Baselines.}
We compared against Base Model (w/o and w/ Context), D2L~\citep{charakorn2026doc}, CD~\citep{snell2022learning}, and AnyEdit~\citep{jiang2025anyedit}, further reporting CD and D2L oracles as reference points.
Detailed baseline configurations are in Appendix~\ref{app:baselines} for reproducibility.

\paragraph{Evaluation Metrics.}
We evaluated performance using ROUGE-L Recall, LLM-as-a-Judge, and Locality, while additionally reporting update and generation latency along with peak memory usage as efficiency metrics, with human evaluation detailed in the Appendix E.
All methods used identical generation settings unless otherwise specified.

\subsection{Main Results}
\label{sec:main_results}

As shown in Table~\ref{tab:main_results}, \method{} \textbf{performed best among practical parameterized methods across all five datasets}, improving average ROUGE-L Recall / LLM-as-a-Judge from 44.46/34.49 to 57.76/53.41 (\textbf{$\sim$29.92\%$\uparrow$}/\textbf{$\sim$54.84\%$\uparrow$}) over D2L.
On SQuAD, \method{} improves ROUGE-L Recall and LLM-as-a-Judge from 54.96/37.27 to \textbf{80.65}/\textbf{76.81} (\textbf{$\sim$46.74\%$\uparrow$}/\textbf{$\sim$106.10\%$\uparrow$}).
Meanwhile, on the longer, more challenging 2Wiki, MFQA, and QASPER datasets, it maintains consistent improvements (e.g., 36.30/30.84 to 47.24/41.48 on 2Wiki, \textbf{$\sim$30.14\%$\uparrow$}/\textbf{$\sim$34.50\%$\uparrow$}).
By pairing a global update representation with query-activated memory evidence, \method{} better preserves currently valid information without providing the full context history to the target model.
Under continual updates, Recall declines by only 2.09 points from 3 to 20 updates (Appendix~\ref{app:multi-update-experiments}).

\subsection{Efficiency and Information Retention}

\label{sec:efficiency_locality}

\paragraph{Computational Efficiency.}
We evaluated update and generation efficiency on SQuAD and 2Wiki in terms of latency and peak memory usage. 
\method{} \textbf{maintained update efficiency on par with lightweight context-parameterization methods while incurring only modest additional overhead during generation}. 
CD and the parameter-editing baselines, in contrast, required substantially greater optimization or update costs across both benchmarks. 
These results collectively demonstrated that \method{} improved robustness to continual updates without sacrificing the efficiency benefits characteristic of reusable parameterized contexts. 
Detailed experimental results are provided in Appendix~\ref{app:efficiency-measurement}.

\begin{figure}[!t]
\centering
\includegraphics[width=\columnwidth]{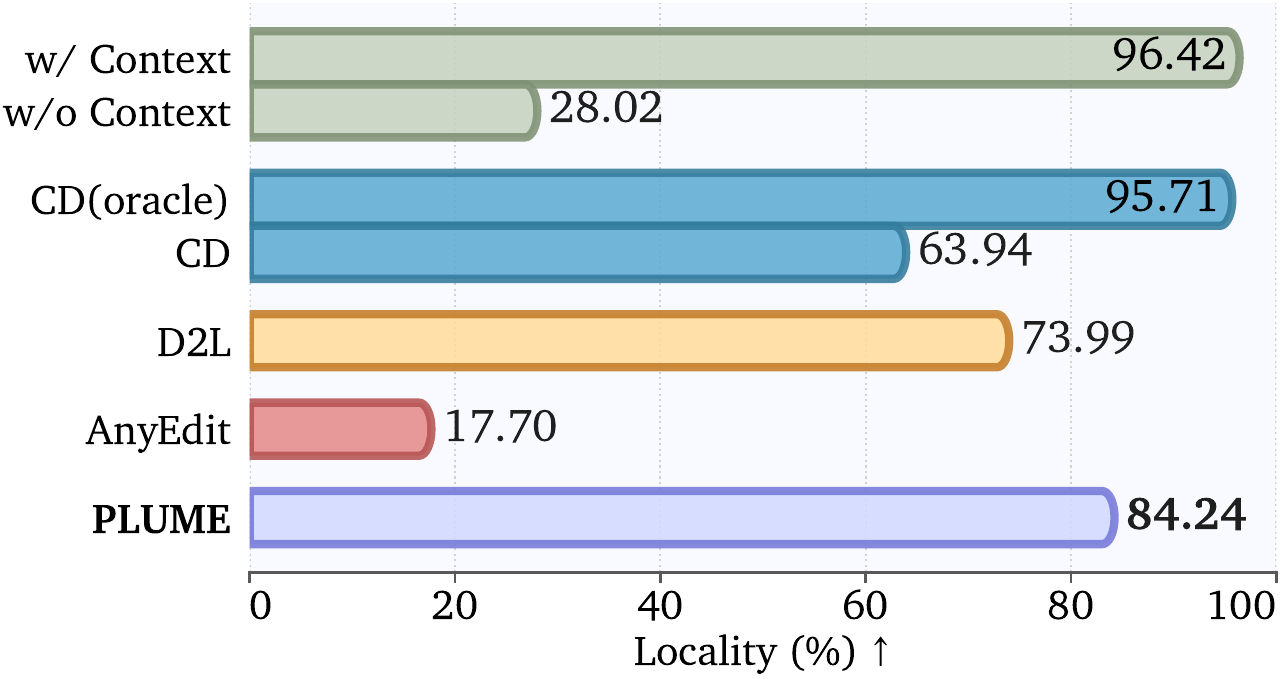}
\captionsetup{skip=3.5pt}
\caption{\textbf{Locality under updates.} Higher scores indicate better unaffected information preservation.}
\label{fig:locality}
\vspace{-4pt} %
\end{figure}

\paragraph{Locality.}
Locality evaluates whether the model preserves answers to queries outside the scope of the update, thereby quantifying interference with previously retained information.
As shown in Figure~\ref{fig:locality}, \method{} \textbf{attains the strongest Locality among parameterized methods}, surpassing D2L and CD by 10.25 (\textbf{$\sim$13.85\%$\uparrow$}) and \textbf{20.30} (\textbf{$\sim$31.75\%$\uparrow$}) points, respectively.
This result indicates that \method{} more effectively confines the interference introduced by parameter updates, thereby better preserving the established context state.

\begin{table}[t]
\centering
\captionsetup{skip=3.5pt}
\footnotesize
\setlength{\tabcolsep}{3.1pt}
\renewcommand{\arraystretch}{1.00}
\setlength{\aboverulesep}{0pt}
\setlength{\belowrulesep}{0pt}
\resizebox{\columnwidth}{!}{%
\begin{tabular}{lrrrr}
\toprule[1.2pt]
\multicolumn{1}{c}{\multirow{2}{*}{\textbf{Method}}} & \multicolumn{2}{c}{\textbf{GSM8K}} & \multicolumn{2}{c}{\textbf{CRUXEval}} \\
\cmidrule(l{0pt}r{1pt}){2-3}\cmidrule(l{1pt}r{0pt}){4-5}
& \textbf{Recall} & \textbf{LLM} & \textbf{Recall} & \textbf{LLM} \\
\midrule[1.2pt]
\multicolumn{5}{c}{\cellcolor{modelshade}\textbf{\llmname{Qwen3}}} \\
w/ Context (oracle) & \textcolor{gray}{54.21} & \textcolor{gray}{52.76} & \textcolor{gray}{68.43} & \textcolor{gray}{55.37} \\
CD (oracle)      & \textcolor{gray}{51.54} & \textcolor{gray}{50.83} & \textcolor{gray}{65.07} & \textcolor{gray}{53.20} \\
D2L (oracle)     & \textcolor{gray}{36.23} & \textcolor{gray}{30.24} & \textcolor{gray}{44.09} & \textcolor{gray}{20.50} \\
\midrule
w/o Context       & 15.21 &  7.82 & 37.75 &  4.62 \\
AnyEdit           & 16.99 &  1.90 & 44.41 & 10.50 \\
CD               & \underline{36.40} & \underline{26.31} & \underline{51.52} & \underline{13.75} \\
D2L              & 23.18 & 16.42 & 39.48 & 13.63 \\
\midrule[0.5pt]
\method{}   & \textbf{45.39} & \textbf{41.47} & \textbf{62.31} & \textbf{50.62} \\
\midrule[1.2pt]
\multicolumn{5}{c}{\cellcolor{modelshade}\textbf{\llmname{Gemma2}}} \\
w/ Context (oracle) & \textcolor{gray}{42.90} & \textcolor{gray}{39.86} & \textcolor{gray}{47.37} & \textcolor{gray}{33.23} \\
CD (oracle)      & \textcolor{gray}{41.96} & \textcolor{gray}{39.92} & \textcolor{gray}{47.09} & \textcolor{gray}{31.81} \\
D2L (oracle)     & \textcolor{gray}{22.23} & \textcolor{gray}{15.29} & \textcolor{gray}{34.36} & \textcolor{gray}{12.83} \\
\midrule
w/o Context       & 10.12 &  4.41 & 18.07 &  2.96 \\
AnyEdit           & 10.67 &  1.26 & \underline{34.35} &  6.77 \\
CD               & \underline{23.03} & \underline{12.52} & 30.55 & 10.74 \\
D2L              & 20.04 & 11.74 & 29.95 & \underline{11.44} \\
\midrule[0.5pt]
\method{}   & \textbf{33.69} & \textbf{29.65} & \textbf{40.52} & \textbf{26.12} \\
\bottomrule[1.2pt]
\end{tabular}
}
\caption{\textbf{Cross-task generalization results.} ROUGE-L Recall and LLM-as-a-Judge are reported.}
\label{tab:cross_task_generalization}
\end{table}

\subsection{Generalization and Overall Comparison}
\label{sec:generalization_overall}

\begin{figure}[!t]
\centering
\includegraphics[width=1\columnwidth]{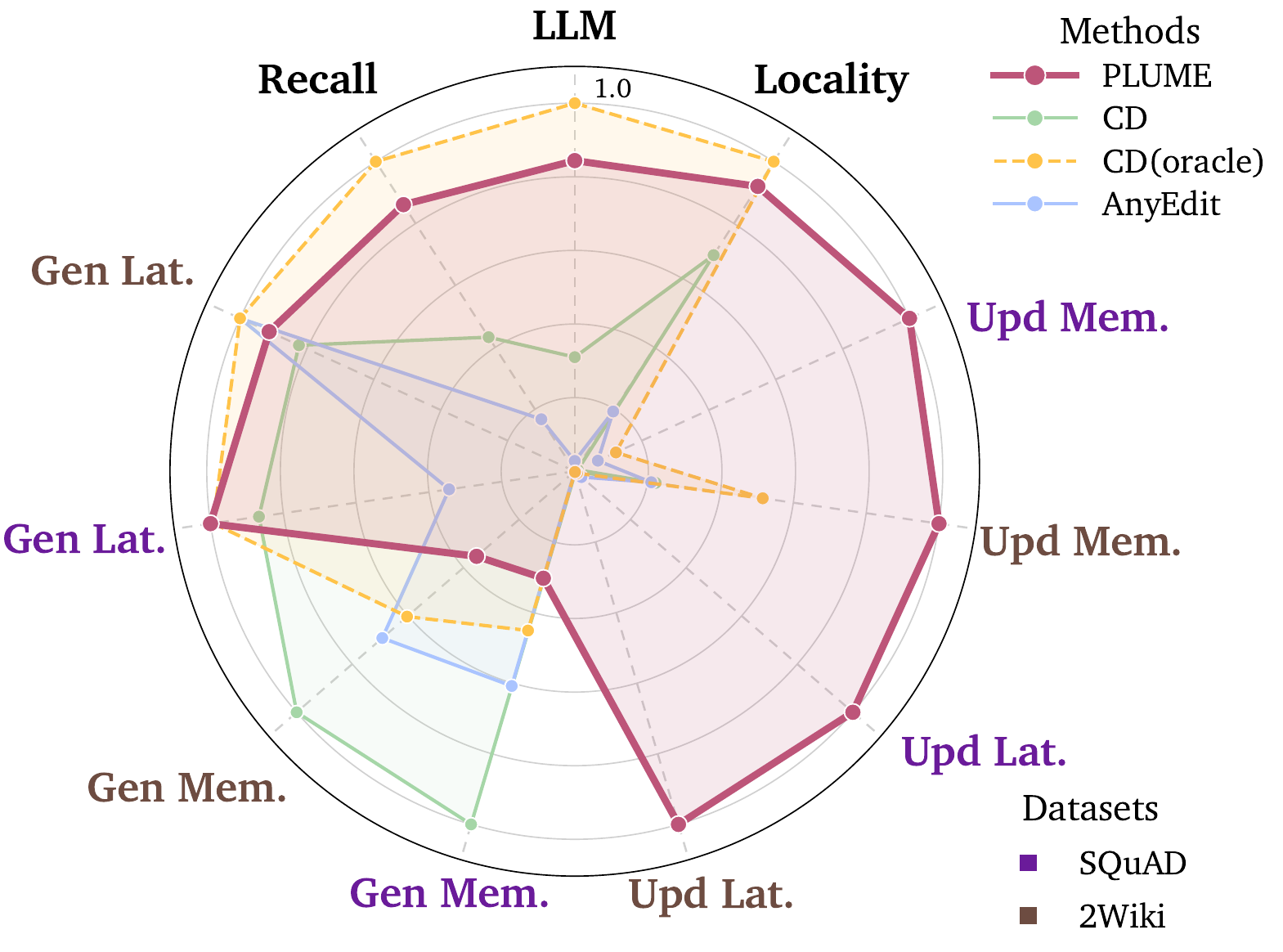}
\captionsetup{skip=3.5pt}
\caption{\textbf{Overall comparison of effectiveness and efficiency.} Normalized metrics include answer quality, locality, and update/generation efficiency.}
\label{fig:overall_radar}
\vspace{-4pt} %
\end{figure}

We further extend the sequential evolution setting to GSM8K and CRUXEval.
As shown in Table~\ref{tab:cross_task_generalization}, \method{} consistently outperforms D2L across tasks:
\ding{172} on GSM8K, the average scores improve from 23.18/16.42 to 45.39/41.47 (\textbf{$\sim$95.82\%$\uparrow$}/\textbf{$\sim$152.56\%$\uparrow$}); 
\ding{173} on CRUXEval, they increase from 39.48/13.63 to 62.31/50.62 (\textbf{$\sim$57.83\%$\uparrow$}/\textbf{$\sim$271.39\%$\uparrow$}).
These results demonstrate that \method{} \textbf{generalizes beyond the QA setting} of \benchmark{}.
Figure~\ref{fig:overall_radar} further shows balanced improvements across evaluation dimensions, together with strong efficiency and robust information retention under continual updates.

\begin{figure}[tbp]
\centering
\includegraphics[width=1\columnwidth]{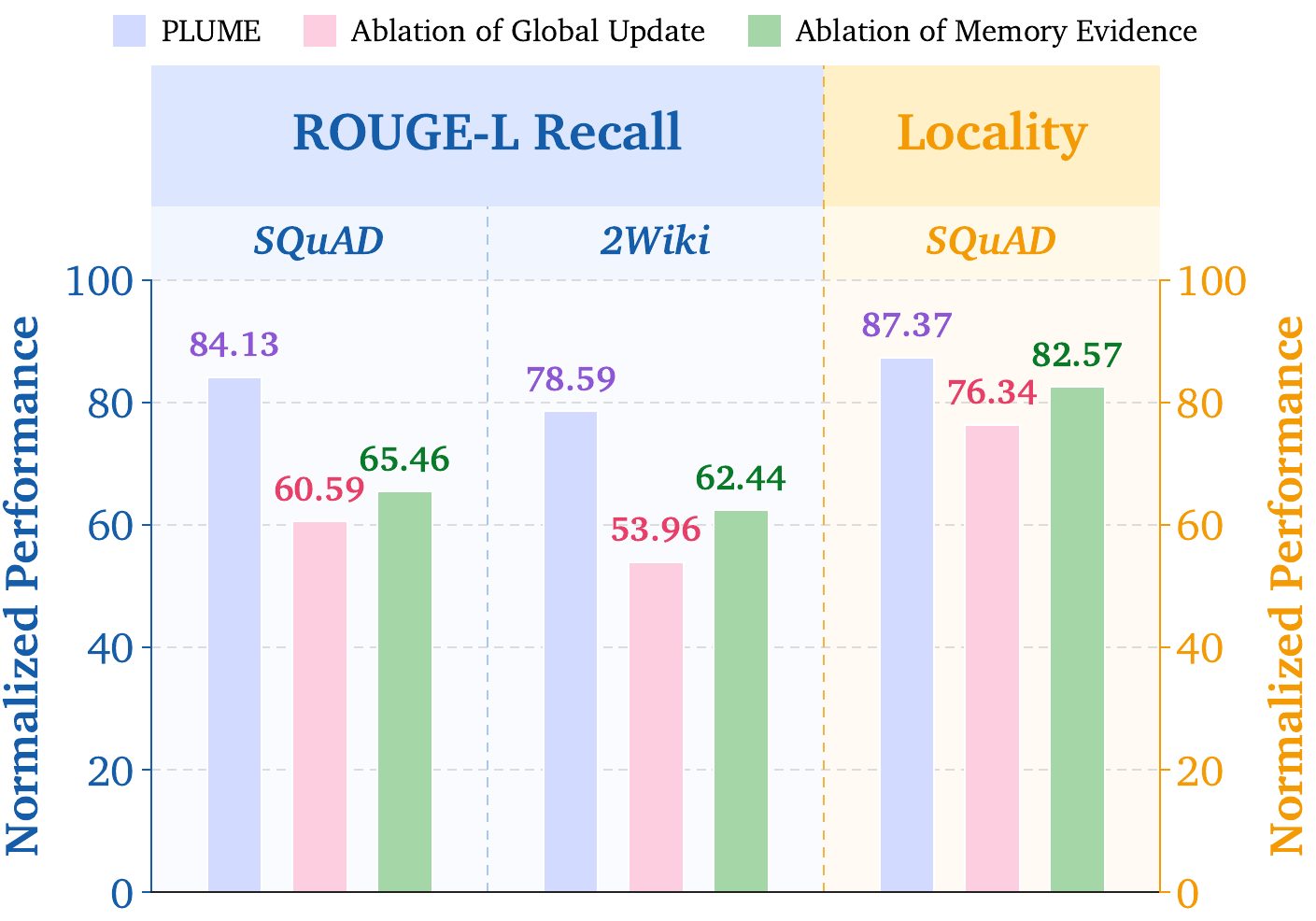}
\captionsetup{skip=3.5pt}
\caption{\textbf{Ablation of \method{} components.} The memory-only and global-only variants each underperform \method{}, demonstrating the complementary contributions of the global update representation and query-activated memory evidence.}
\label{fig:ablation}
\end{figure}

\begin{figure*}[!t]
    \centering
    \includegraphics[width=\textwidth]{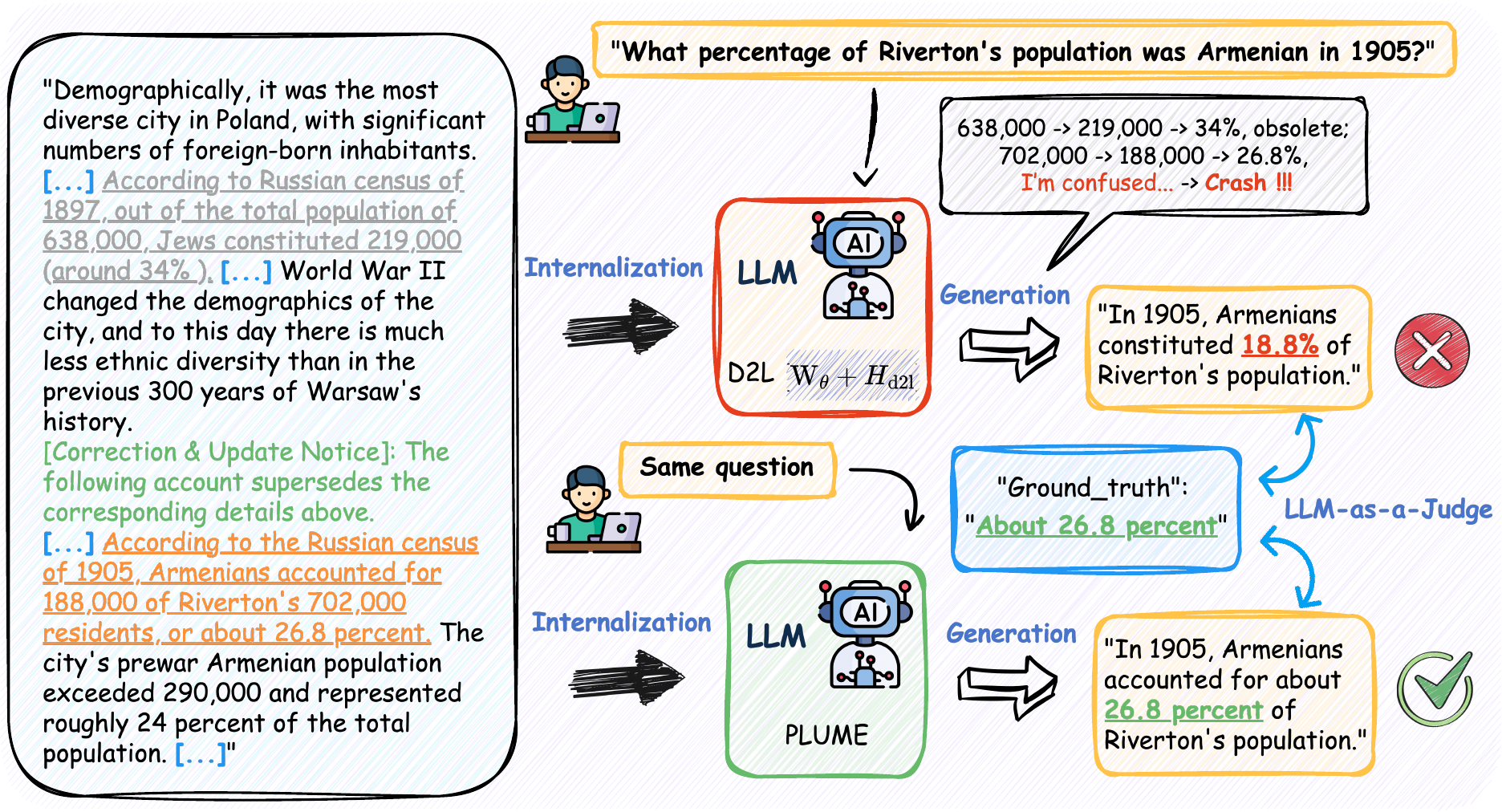}
    \caption{\textbf{Representative case under a context update.} D2L conflates information with different validity states and produces an incorrect answer, whereas \method{} follows the currently valid update.}
    \label{fig:case_study}
\end{figure*}

\subsection{Ablation Study}
\label{sec:ablation}
We compared \method{} with two component ablations: a memory-only variant and a global-only variant.
As shown in Figure~\ref{fig:ablation}, \method{} achieved an average normalized ROUGE-L Recall score of \textbf{81.36}, compared with 57.28 \textit{without the global update representation} and 63.95 \textit{without memory evidence}, corresponding to improvements of \textbf{24.09} (\textbf{$\sim$42.05\%$\uparrow$}) and \textbf{17.41} (\textbf{$\sim$27.22\%$\uparrow$}) points.
For normalized Locality, \method{} reached \textbf{87.37}, exceeding the two ablations by \textbf{11.03} (\textbf{$\sim$14.45\%$\uparrow$}) and \textbf{4.80} (\textbf{$\sim$5.81\%$\uparrow$}) points.
\textbf{The larger degradation without the global update representation highlights its role in maintaining the current context state}, while memory evidence provides complementary gains in answer quality and preserving unaffected information.

\subsection{Case Study}
\label{sec:case_study}
Representative cases contained valid, invalidated, and unaffected information within the same parameterized state, creating a substantial challenge for reliable validity-state disambiguation.
As shown in Figure~\ref{fig:case_study}, D2L became confused by competing information with different validity states and generated an incorrect answer, whereas \method{} better captured the global update representation and emphasized the latest valid evidence.
This behavior is consistent with the validity-state interference observed in the aggregate results and further reinforces our quantitative findings.

\section{Related Work}
\label{sec:related-work}

\paragraph{\rpsStrongTerm{Context Internalization.}}

RAG~\citep{gao2023retrieval} and long-context models~\citep{ding2023longnet} supply context at inference time, incurring repeated overhead.
Context parameterization instead internalizes context into reusable representations.
CD transfers context-induced behavior into model parameters~\citep{snell2022learning}, while D2L compiles context into lightweight parameterized representations~\citep{charakorn2026doc}.
Both, however, presuppose static, internally consistent contexts, leaving validity changes under updates unaddressed.

\paragraph{\rpsStrongTerm{Knowledge Editing and Continual Updates.}}
Knowledge editing modifies factual knowledge in language models without full retraining~\citep{yao2023editing}, with representative methods including ROME, MEMIT, and AlphaEdit~\citep{meng2022locating,meng2023massediting,fang2025alphaedit}.
Recent work has further examined sequential editing, interference, and reliable preservation of prior knowledge~\citep{wang2024knowledge,chen2024lifelong,jiang2024learning,li2025reinforced}.
Our setting instead studies continual updates to parameterized contexts with factual dependencies and unaffected information.

\paragraph{\rpsStrongTerm{Knowledge Conflicts.}}
Prior work has primarily examined conflicts between parametric knowledge and external context~\citep{xie2024adaptive,shi2024ircan}, or conflicts among retrieved contexts~\citep{cattan2025dragged,xu2024knowledge}.
These settings expose conflicting evidence directly to the model at inference time.
In contrast, our setting concerns currently valid and invalidated information already jointly internalized within parameters, without direct access to the original context. \benchmark{} and \method{} are designed for this setting.

\section{Conclusion}
\label{sec:conclusion}
We introduced \task{} and \benchmark{} to study context parameterization under continual updates, jointly evaluating update incorporation and preservation of unaffected knowledge.
Our training-free method, \method{}, combines a global update representation with query-activated memory evidence through adaptive decoding.
Experiments across datasets and backbones demonstrated improvements in answer quality and locality, with further gains on mathematical and code reasoning tasks.
Ablations supported the complementary roles of global representations and local evidence.
These findings support coordinating global updates with selective evidence access to build reliable, reusable parameterized memory.

\section*{Limitations}
\label{sec:limitations}

While we believe that \method{} offers a valuable contribution to context parameterization and utilization for continually updated knowledge, several limitations should be acknowledged. First, \method{} is built entirely atop D2L, whose hypernetwork compresses a context into a compact set of LoRA parameters; \textbf{this parameterization is inherently lossy}, and long or information-dense contexts may not be faithfully preserved during encoding, which in turn upper-bounds the overall performance of \method{} by the fidelity and capacity of the underlying parameterization method. Second, \method{} is deliberately training-free, a design choice that preserves its flexibility, low memory footprint, and low latency, but that also confines its interventions to parameter construction and decoding time, \textbf{without directly enhancing the hypernetwork's intrinsic ability to discriminate valid updates from invalidated information}. Our future work will therefore focus on jointly improving the fidelity of context parameterization and equipping the hypernetwork itself with update-aware, training-based mechanisms, aiming to provide continued valuable insights for the development of truly reliable context-adaptive systems.
\flushbottom

\bibliography{main}

\begin{thebibliography}{50}
\providecommand{\natexlab}[1]{#1}

\bibitem[{Askell et~al.(2021)Askell, Bai, Chen, Drain, Ganguli, Henighan,
  Jones, Joseph, Mann, DasSarma et~al.}]{askell2021general}
Amanda Askell, Yuntao Bai, Anna Chen, Dawn Drain, Deep Ganguli, Tom Henighan,
  Andy Jones, Nicholas Joseph, Ben Mann, Nova DasSarma, and 1 others. 2021.
\newblock A general language assistant as a laboratory for alignment.
\newblock \emph{arXiv preprint arXiv:2112.00861}.

\bibitem[{Bai et~al.(2024)Bai, Lv, Zhang, Lyu, Tang, Huang, Du, Liu, Zeng, Hou
  et~al.}]{bai2024longbench}
Yushi Bai, Xin Lv, Jiajie Zhang, Hongchang Lyu, Jiankai Tang, Zhidian Huang,
  Zhengxiao Du, Xiao Liu, Aohan Zeng, Lei Hou, and 1 others. 2024.
\newblock Longbench: A bilingual, multitask benchmark for long context
  understanding.
\newblock In \emph{Proceedings of the 62nd annual meeting of the association
  for computational linguistics (volume 1: Long papers)}, pages 3119--3137.

\bibitem[{Bai et~al.(2025)Bai, Tu, Zhang, Peng, Wang, Lv, Cao, Xu, Hou, Dong
  et~al.}]{bai2025longbench}
Yushi Bai, Shangqing Tu, Jiajie Zhang, Hao Peng, Xiaozhi Wang, Xin Lv, Shulin
  Cao, Jiazheng Xu, Lei Hou, Yuxiao Dong, and 1 others. 2025.
\newblock Longbench v2: Towards deeper understanding and reasoning on realistic
  long-context multitasks.
\newblock In \emph{Proceedings of the 63rd Annual Meeting of the Association
  for Computational Linguistics (Volume 1: Long Papers)}, pages 3639--3664.

\bibitem[{Brown et~al.(2020)Brown, Mann, Ryder, Subbiah, Kaplan, Dhariwal,
  Neelakantan, Shyam, Sastry, Askell et~al.}]{brown2020language}
Tom Brown, Benjamin Mann, Nick Ryder, Melanie Subbiah, Jared~D Kaplan, Prafulla
  Dhariwal, Arvind Neelakantan, Pranav Shyam, Girish Sastry, Amanda Askell, and
  1 others. 2020.
\newblock Language models are few-shot learners.
\newblock \emph{Advances in neural information processing systems},
  33:1877--1901.

\bibitem[{Cattan et~al.(2025)Cattan, Jacovi, Ram, Herzig, Aharoni, Goldshtein,
  Ofek, Szpektor, and Caciularu}]{cattan2025dragged}
Arie Cattan, Alon Jacovi, Ori Ram, Jonathan Herzig, Roee Aharoni, Sasha
  Goldshtein, Eran Ofek, Idan Szpektor, and Avi Caciularu. 2025.
\newblock Dragged into conflicts: Detecting and addressing conflicting sources
  in search-augmented llms.
\newblock \emph{arXiv preprint arXiv:2506.08500}.

\bibitem[{Charakorn et~al.(2026)Charakorn, Cetin, Uesaka, and
  Lange}]{charakorn2026doc}
Rujikorn Charakorn, Edoardo Cetin, Shinnosuke Uesaka, and Robert~Tjarko Lange.
  2026.
\newblock Doc-to-lora: Learning to instantly internalize contexts.
\newblock \emph{arXiv preprint arXiv:2602.15902}.

\bibitem[{Chen et~al.(2024)Chen, Zhang, He, Li, Wang, Huang
  et~al.}]{chen2024lifelong}
Qizhou Chen, Taolin Zhang, Xiaofeng He, Dongyang Li, Chengyu Wang, Longtao
  Huang, and 1 others. 2024.
\newblock Lifelong knowledge editing for llms with retrieval-augmented
  continuous prompt learning.
\newblock In \emph{Proceedings of the 2024 conference on empirical methods in
  natural language processing}, pages 13565--13580.

\bibitem[{Chen et~al.(2023)Chen, Wong, Chen, and Tian}]{chen2023extending}
Shouyuan Chen, Sherman Wong, Liangjian Chen, and Yuandong Tian. 2023.
\newblock Extending context window of large language models via positional
  interpolation.
\newblock \emph{arXiv preprint arXiv:2306.15595}.

\bibitem[{Cobbe et~al.(2021)Cobbe, Kosaraju, Bavarian, Chen, Jun, Kaiser,
  Plappert, Tworek, Hilton, Nakano et~al.}]{cobbe2021training}
Karl Cobbe, Vineet Kosaraju, Mohammad Bavarian, Mark Chen, Heewoo Jun, Lukasz
  Kaiser, Matthias Plappert, Jerry Tworek, Jacob Hilton, Reiichiro Nakano, and
  1 others. 2021.
\newblock Training verifiers to solve math word problems.
\newblock \emph{arXiv preprint arXiv:2110.14168}.

\bibitem[{Dasigi et~al.(2021)Dasigi, Lo, Beltagy, Cohan, Smith, and
  Gardner}]{dasigi2021dataset}
Pradeep Dasigi, Kyle Lo, Iz~Beltagy, Arman Cohan, Noah~A Smith, and Matt
  Gardner. 2021.
\newblock A dataset of information-seeking questions and answers anchored in
  research papers.
\newblock In \emph{Proceedings of the 2021 Conference of the North American
  Chapter of the Association for Computational Linguistics: Human Language
  Technologies}, pages 4599--4610.

\bibitem[{Ding et~al.(2023)Ding, Ma, Dong, Zhang, Huang, Wang, Zheng, and
  Wei}]{ding2023longnet}
Jiayu Ding, Shuming Ma, Li~Dong, Xingxing Zhang, Shaohan Huang, Wenhui Wang,
  Nanning Zheng, and Furu Wei. 2023.
\newblock Longnet: Scaling transformers to 1,000,000,000 tokens.
\newblock \emph{arXiv preprint arXiv:2307.02486}.

\bibitem[{Fang et~al.(2025)Fang, Jiang, Wang, Ma, Shi, Wang, He, and
  Chua}]{fang2025alphaedit}
Junfeng Fang, Houcheng Jiang, Kun Wang, Yunshan Ma, Jie Shi, Xiang Wang,
  Xiangnan He, and Tat-Seng Chua. 2025.
\newblock Alphaedit: Null-space constrained knowledge editing for language
  models.
\newblock In \emph{International Conference on Learning Representations},
  volume 2025, pages 16366--16396.

\bibitem[{Gao et~al.(2023)Gao, Xiong, Gao, Jia, Pan, Bi, Dai, Sun, Wang, and
  Wang}]{gao2023retrieval}
Yunfan Gao, Yun Xiong, Xinyu Gao, Kangxiang Jia, Jinliu Pan, Yuxi Bi, Yi~Dai,
  Jiawei Sun, Meng Wang, and Haofen Wang. 2023.
\newblock Retrieval-augmented generation for large language models: A survey.
\newblock \emph{arXiv preprint arXiv:2312.10997}.

\bibitem[{Gu et~al.(2024)Gu, Rozi{\`e}re, Leather, Solar-Lezama, Synnaeve, and
  Wang}]{gu2024cruxeval}
Alex Gu, Baptiste Rozi{\`e}re, Hugh Leather, Armando Solar-Lezama, Gabriel
  Synnaeve, and Sida~I Wang. 2024.
\newblock Cruxeval: A benchmark for code reasoning, understanding and
  execution.
\newblock \emph{arXiv preprint arXiv:2401.03065}.

\bibitem[{Guo et~al.(2025)Guo, Yang, Zhang, Song, Wang, Zhu, Xu, Zhang, Ma, Bi
  et~al.}]{guo2025deepseek}
Daya Guo, Dejian Yang, Haowei Zhang, Junxiao Song, Peiyi Wang, Qihao Zhu,
  Runxin Xu, Ruoyu Zhang, Shirong Ma, Xiao Bi, and 1 others. 2025.
\newblock Deepseek-r1: Incentivizing reasoning capability in llms via
  reinforcement learning.
\newblock \emph{arXiv preprint arXiv:2501.12948}.

\bibitem[{Guu et~al.(2020)Guu, Lee, Tung, Pasupat, and
  Chang}]{guu2020retrieval}
Kelvin Guu, Kenton Lee, Zora Tung, Panupong Pasupat, and Mingwei Chang. 2020.
\newblock Retrieval augmented language model pre-training.
\newblock In \emph{International conference on machine learning}, pages
  3929--3938. PMLR.

\bibitem[{Ho et~al.(2020)Ho, Nguyen, Sugawara, and Aizawa}]{ho2020constructing}
Xanh Ho, Anh-Khoa~Duong Nguyen, Saku Sugawara, and Akiko Aizawa. 2020.
\newblock Constructing a multi-hop qa dataset for comprehensive evaluation of
  reasoning steps.
\newblock In \emph{Proceedings of the 28th International Conference on
  Computational Linguistics}, pages 6609--6625.

\bibitem[{Hsieh et~al.(2024)Hsieh, Chuang, Li, Wang, Le, Kumar, Glass, Ratner,
  Lee, Krishna et~al.}]{hsieh2024found}
Cheng-Yu Hsieh, Yung-Sung Chuang, Chun-Liang Li, Zifeng Wang, Long Le, Abhishek
  Kumar, James Glass, Alexander Ratner, Chen-Yu Lee, Ranjay Krishna, and 1
  others. 2024.
\newblock Found in the middle: Calibrating positional attention bias improves
  long context utilization.
\newblock In \emph{Findings of the Association for Computational Linguistics:
  ACL 2024}, pages 14982--14995.

\bibitem[{Hu et~al.(2022)Hu, yelong shen, Wallis, Allen-Zhu, Li, Wang, Wang,
  and Chen}]{hu2022lora}
Edward~J Hu, yelong shen, Phillip Wallis, Zeyuan Allen-Zhu, Yuanzhi Li, Shean
  Wang, Lu~Wang, and Weizhu Chen. 2022.
\newblock \href {https://openreview.net/forum?id=nZeVKeeFYf9} {Lo{RA}: Low-rank
  adaptation of large language models}.
\newblock In \emph{International Conference on Learning Representations}.

\bibitem[{Jang et~al.(2022)Jang, Ye, Lee, Yang, Shin, Han, Kim, and
  Seo}]{jang2022temporalwiki}
Joel Jang, Seonghyeon Ye, Changho Lee, Sohee Yang, Joongbo Shin, Janghoon Han,
  Gyeonghun Kim, and Minjoon Seo. 2022.
\newblock Temporalwiki: A lifelong benchmark for training and evaluating
  ever-evolving language models.
\newblock In \emph{Proceedings of the 2022 Conference on Empirical Methods in
  Natural Language Processing}, pages 6237--6250.

\bibitem[{Jiang et~al.(2025)Jiang, Fang, Zhang, Wan, Ma, Wang, He, and
  Chua}]{jiang2025anyedit}
Houcheng Jiang, Junfeng Fang, Ningyu Zhang, Mingyang Wan, Guojun Ma, Xiang
  Wang, Xiangnan He, and Tat-Seng Chua. 2025.
\newblock \href {https://openreview.net/forum?id=aJIoBur0Ef} {Anyedit: Edit any
  knowledge encoded in language models}.
\newblock In \emph{Forty-second International Conference on Machine Learning}.

\bibitem[{Jiang et~al.(2024)Jiang, Wang, Wu, Zhong, Zeng, Gao, Li, Jiang,
  Shang, Tang et~al.}]{jiang2024learning}
Yuxin Jiang, Yufei Wang, Chuhan Wu, Wanjun Zhong, Xingshan Zeng, Jiahui Gao,
  Liangyou Li, Xin Jiang, Lifeng Shang, Ruiming Tang, and 1 others. 2024.
\newblock Learning to edit: Aligning llms with knowledge editing.
\newblock In \emph{Proceedings of the 62nd Annual Meeting of the Association
  for Computational Linguistics (Volume 1: Long Papers)}, pages 4689--4705.

\bibitem[{Kortukov et~al.(2024)Kortukov, Rubinstein, Nguyen, and
  Oh}]{kortukov2024studying}
Evgenii Kortukov, Alexander Rubinstein, Elisa Nguyen, and Seong~Joon Oh. 2024.
\newblock \href {https://openreview.net/forum?id=xm8zYRfrqE} {Studying large
  language model behaviors under context-memory conflicts with real documents}.
\newblock In \emph{First Conference on Language Modeling}.

\bibitem[{Lewis et~al.(2020)Lewis, Perez, Piktus, Petroni, Karpukhin, Goyal,
  K{\"u}ttler, Lewis, Yih, Rockt{\"a}schel et~al.}]{lewis2020retrieval}
Patrick Lewis, Ethan Perez, Aleksandra Piktus, Fabio Petroni, Vladimir
  Karpukhin, Naman Goyal, Heinrich K{\"u}ttler, Mike Lewis, Wen-tau Yih, Tim
  Rockt{\"a}schel, and 1 others. 2020.
\newblock Retrieval-augmented generation for knowledge-intensive nlp tasks.
\newblock \emph{Advances in neural information processing systems},
  33:9459--9474.

\bibitem[{Li et~al.(2025{\natexlab{a}})Li, Jiang, Wu, Luo, Ahn, Zhang, Abdi,
  Li, Gao, Yang et~al.}]{li2025scbench}
Yucheng Li, Huiqiang Jiang, Qianhui Wu, Xufang Luo, Surin Ahn, Chengruidong
  Zhang, Amir Abdi, Dongsheng Li, Jianfeng Gao, Yuqing Yang, and 1 others.
  2025{\natexlab{a}}.
\newblock Scbench: A kv cache-centric analysis of long-context methods.
\newblock In \emph{International Conference on Learning Representations},
  volume 2025, pages 66063--66093.

\bibitem[{Li et~al.(2025{\natexlab{b}})Li, Jiang, Chen, Bi, Zhou, Sun, Fang,
  and Wang}]{li2025reinforced}
Zherui Li, Houcheng Jiang, Hao Chen, Baolong Bi, Zhenhong Zhou, Fei Sun,
  Junfeng Fang, and Xiang Wang. 2025{\natexlab{b}}.
\newblock \href {https://openreview.net/forum?id=1jUXprrfcb} {Reinforced
  lifelong editing for language models}.
\newblock In \emph{Forty-second International Conference on Machine Learning}.

\bibitem[{Lin(2004)}]{lin2004rouge}
Chin-Yew Lin. 2004.
\newblock Rouge: A package for automatic evaluation of summaries.
\newblock In \emph{Text summarization branches out}, pages 74--81.

\bibitem[{Lin et~al.(2019)Lin, Tafjord, Clark, and Gardner}]{lin2019reasoning}
Kevin Lin, Oyvind Tafjord, Peter Clark, and Matt Gardner. 2019.
\newblock Reasoning over paragraph effects in situations.
\newblock In \emph{Proceedings of the 2nd Workshop on Machine Reading for
  Question Answering}, pages 58--62.

\bibitem[{Liu et~al.(2024)Liu, Lin, Hewitt, Paranjape, Bevilacqua, Petroni, and
  Liang}]{liu2024lost}
Nelson~F Liu, Kevin Lin, John Hewitt, Ashwin Paranjape, Michele Bevilacqua,
  Fabio Petroni, and Percy Liang. 2024.
\newblock Lost in the middle: How language models use long contexts.
\newblock \emph{Transactions of the association for computational linguistics},
  12:157--173.

\bibitem[{Liu et~al.(2025)Liu, Xu, Liu, Deng, Wang, Wang, Li, Teh, and
  Lee}]{liu2025model}
Wei Liu, Haomei Xu, Bingqing Liu, Zhiying Deng, Haozhao Wang, Jun Wang, Ruixuan
  Li, Yee~Whye Teh, and Wee~Sun Lee. 2025.
\newblock Is model editing built on sand? revealing its illusory success and
  fragile foundation.
\newblock \emph{arXiv preprint arXiv:2510.00625}.

\bibitem[{Marjanovi{\'c} et~al.(2024)Marjanovi{\'c}, Yu, Atanasova, Maistro,
  Lioma, and Augenstein}]{marjanovic2024dynamicqa}
Sara~Vera Marjanovi{\'c}, Haeun Yu, Pepa Atanasova, Maria Maistro, Christina
  Lioma, and Isabelle Augenstein. 2024.
\newblock Dynamicqa: Tracing internal knowledge conflicts in language models.
\newblock In \emph{Findings of the Association for Computational Linguistics:
  EMNLP 2024}, pages 14346--14360.

\bibitem[{Meng et~al.(2022)Meng, Bau, Andonian, and
  Belinkov}]{meng2022locating}
Kevin Meng, David Bau, Alex Andonian, and Yonatan Belinkov. 2022.
\newblock Locating and editing factual associations in gpt.
\newblock \emph{Advances in neural information processing systems},
  35:17359--17372.

\bibitem[{Meng et~al.(2023)Meng, Sharma, Andonian, Belinkov, and
  Bau}]{meng2023massediting}
Kevin Meng, Arnab~Sen Sharma, Alex~J Andonian, Yonatan Belinkov, and David Bau.
  2023.
\newblock \href {https://openreview.net/forum?id=MkbcAHIYgyS} {Mass-editing
  memory in a transformer}.
\newblock In \emph{The Eleventh International Conference on Learning
  Representations}.

\bibitem[{Mitchell et~al.(2022)Mitchell, Lin, Bosselut, Finn, and
  Manning}]{mitchell2022fast}
Eric Mitchell, Charles Lin, Antoine Bosselut, Chelsea Finn, and Christopher~D
  Manning. 2022.
\newblock \href {https://openreview.net/forum?id=0DcZxeWfOPt} {Fast model
  editing at scale}.
\newblock In \emph{International Conference on Learning Representations}.

\bibitem[{{OpenAI}(2026)}]{openai2026gpt55}
{OpenAI}. 2026.
\newblock {GPT-5.5} system card.
\newblock \url{https://openai.com/index/gpt-5-5-system-card/}.
\newblock Published April 23, 2026; updated April 24, 2026.

\bibitem[{Packer et~al.(2023)Packer, Wooders, Lin, Fang, Patil, Stoica, and
  Gonzalez}]{packer2023memgpt}
Charles Packer, Sarah Wooders, Kevin Lin, Vivian Fang, Shishir~G Patil, Ion
  Stoica, and Joseph~E Gonzalez. 2023.
\newblock Memgpt: Towards llms as operating systems, 2024.
\newblock \emph{URL https://arxiv. org/abs/2310.08560}, 7.

\bibitem[{Park et~al.(2023)Park, O'Brien, Cai, Morris, Liang, and
  Bernstein}]{park2023generative}
Joon~Sung Park, Joseph O'Brien, Carrie~Jun Cai, Meredith~Ringel Morris, Percy
  Liang, and Michael~S Bernstein. 2023.
\newblock Generative agents: Interactive simulacra of human behavior.
\newblock In \emph{Proceedings of the 36th annual acm symposium on user
  interface software and technology}, pages 1--22.

\bibitem[{Rajpurkar et~al.(2016)Rajpurkar, Zhang, Lopyrev, and
  Liang}]{rajpurkar2016squad}
Pranav Rajpurkar, Jian Zhang, Konstantin Lopyrev, and Percy Liang. 2016.
\newblock Squad: 100,000+ questions for machine comprehension of text.
\newblock In \emph{Proceedings of the 2016 conference on empirical methods in
  natural language processing}, pages 2383--2392.

\bibitem[{Shi et~al.(2024)Shi, Jin, Shen, Dong, Wu, and Xiong}]{shi2024ircan}
Dan Shi, Renren Jin, Tianhao Shen, Weilong Dong, Xinwei Wu, and Deyi Xiong.
  2024.
\newblock Ircan: Mitigating knowledge conflicts in llm generation via
  identifying and reweighting context-aware neurons.
\newblock \emph{Advances in Neural Information Processing Systems},
  37:4997--5024.

\bibitem[{Snell et~al.(2022)Snell, Klein, and Zhong}]{snell2022learning}
Charlie Snell, Dan Klein, and Ruiqi Zhong. 2022.
\newblock Learning by distilling context.
\newblock \emph{arXiv preprint arXiv:2209.15189}.

\bibitem[{Team et~al.(2024)Team, Riviere, Pathak, Sessa, Hardin, Bhupatiraju,
  Hussenot, Mesnard, Shahriari, Ram{\'e} et~al.}]{team2024gemma}
Gemma Team, Morgane Riviere, Shreya Pathak, Pier~Giuseppe Sessa, Cassidy
  Hardin, Surya Bhupatiraju, L{\'e}onard Hussenot, Thomas Mesnard, Bobak
  Shahriari, Alexandre Ram{\'e}, and 1 others. 2024.
\newblock Gemma 2: Improving open language models at a practical size.
\newblock \emph{arXiv preprint arXiv:2408.00118}.

\bibitem[{Tian et~al.(2026)Tian, He, Wu, Wang, Chen, Li, and
  Yue}]{tian2026anyedit}
Bowen Tian, Caixue He, Jiemin Wu, Jingying Wang, Wenshuo Chen, Zexi Li, and
  Yutao Yue. 2026.
\newblock \href {https://openreview.net/forum?id=W6qfbvysDh} {Anyedit++:
  Adaptive long-form knowledge editing via bayesian surprise}.
\newblock In \emph{Forty-third International Conference on Machine Learning}.

\bibitem[{Wallat et~al.(2026)Wallat, Nejdl, and Sikdar}]{wallat2026facts}
Jonas Wallat, Wolfgang Nejdl, and Sandipan Sikdar. 2026.
\newblock When facts change: Temporal knowledge conflict resolution in llms.
\newblock In \emph{Findings of the Association for Computational Linguistics:
  ACL 2026}, pages 2154--2184.

\bibitem[{Wang et~al.(2024)Wang, Zhu, Liu, Zheng, Chen, and
  Li}]{wang2024knowledge}
Song Wang, Yaochen Zhu, Haochen Liu, Zaiyi Zheng, Chen Chen, and Jundong Li.
  2024.
\newblock Knowledge editing for large language models: A survey.
\newblock \emph{ACM Computing Surveys}, 57(3):1--37.

\bibitem[{Wei et~al.(2022)Wei, Wang, Schuurmans, Bosma, Xia, Chi, Le, Zhou
  et~al.}]{wei2022chain}
Jason Wei, Xuezhi Wang, Dale Schuurmans, Maarten Bosma, Fei Xia, Ed~Chi, Quoc~V
  Le, Denny Zhou, and 1 others. 2022.
\newblock Chain-of-thought prompting elicits reasoning in large language
  models.
\newblock \emph{Advances in neural information processing systems},
  35:24824--24837.

\bibitem[{Xie et~al.(2024)Xie, Zhang, Chen, Lou, and Su}]{xie2024adaptive}
Jian Xie, Kai Zhang, Jiangjie Chen, Renze Lou, and Yu~Su. 2024.
\newblock Adaptive chameleon or stubborn sloth: Revealing the behavior of large
  language models in knowledge conflicts.
\newblock In \emph{International Conference on Learning Representations},
  volume 2024, pages 35623--35646.

\bibitem[{Xu et~al.(2024)Xu, Qi, Guo, Wang, Wang, Zhang, and
  Xu}]{xu2024knowledge}
Rongwu Xu, Zehan Qi, Zhijiang Guo, Cunxiang Wang, Hongru Wang, Yue Zhang, and
  Wei Xu. 2024.
\newblock Knowledge conflicts for llms: A survey.
\newblock In \emph{Proceedings of the 2024 Conference on Empirical Methods in
  Natural Language Processing}, pages 8541--8565.

\bibitem[{Yang et~al.(2025)Yang, Li, Yang, Zhang, Hui, Zheng, Yu, Gao, Huang,
  Lv et~al.}]{yang2025qwen3}
An~Yang, Anfeng Li, Baosong Yang, Beichen Zhang, Binyuan Hui, Bo~Zheng, Bowen
  Yu, Chang Gao, Chengen Huang, Chenxu Lv, and 1 others. 2025.
\newblock Qwen3 technical report.
\newblock \emph{arXiv preprint arXiv:2505.09388}.

\bibitem[{Yao et~al.(2023)Yao, Wang, Tian, Cheng, Li, Deng, Chen, and
  Zhang}]{yao2023editing}
Yunzhi Yao, Peng Wang, Bozhong Tian, Siyuan Cheng, Zhoubo Li, Shumin Deng,
  Huajun Chen, and Ningyu Zhang. 2023.
\newblock Editing large language models: Problems, methods, and opportunities.
\newblock In \emph{Proceedings of the 2023 Conference on Empirical Methods in
  Natural Language Processing}, pages 10222--10240.

\bibitem[{Zhong et~al.(2024)Zhong, Guo, Gao, Ye, and
  Wang}]{zhong2024memorybank}
Wanjun Zhong, Lianghong Guo, Qiqi Gao, He~Ye, and Yanlin Wang. 2024.
\newblock Memorybank: Enhancing large language models with long-term memory.
\newblock In \emph{Proceedings of the AAAI conference on artificial
  intelligence}, volume~38, pages 19724--19731.

\end{thebibliography}

\clearpage
\appendix
\section{LLM Usage Statement}
We utilized Large Language Models to assist with the writing and polishing of the manuscript, including improvements to grammar, clarity, conciseness, and word choice. As described in Section~\ref{sec:benchmark}, LLMs were also used for data construction, quality verification, and automatic evaluation. However, they were not used to formulate the core research ideas or design the proposed methodology.

\section{Detailed Setup}
\label{app:detailed-setup}
In this section, we provide additional details regarding the models, benchmark, datasets, and baseline methods used in our experiments.

\subsection{Models}
\label{app:models}
We used two base language models and their corresponding pretrained D2L hypernetworks:
\begin{itemize}
    \item \textbf{\llmname{Qwen3-4B-Instruct-2507}.}
    \llmname{Qwen3-4B-Instruct-2507} is a 4B-parameter instruction-tuned autoregressive language model from the \llmname{Qwen3} family~\citep{yang2025qwen3}. It served as our primary backbone and provided a relatively strong instruction-following capability at a moderate model scale. We downloaded the \nolinkurl{Qwen/Qwen3-4B-Instruct-2507} checkpoint from Hugging Face.

    \item \textbf{\llmname{Gemma-2-2B-it}.}
    \llmname{Gemma-2-2B-it} is a 2B-parameter instruction-tuned autoregressive language model from the \llmname{Gemma 2} family~\citep{team2024gemma}. Compared with \llmname{Qwen3-4B-Instruct-2507}, it provided a smaller and architecturally distinct backbone for evaluating the robustness of our method across model families and scales. We downloaded the \texttt{google/gemma-2-2b-it} checkpoint from Hugging Face.

    \item \textbf{\llmname{Qwen-4B} D2L hypernetwork.} The \llmname{Qwen-4B} D2L hypernetwork is a pretrained context-to-parameter model that maps textual context directly into LoRA parameters conditioned on the target backbone, \llmname{Qwen3-4B-Instruct-2507}~\citep{charakorn2026doc}.
    It enables long-form context information to be compressed into a lightweight parametric representation without directly modifying the frozen backbone parameters.
    We used the \nolinkurl{qwen_4b_d2l/checkpoint-20000} checkpoint after 20K training steps.

    \item \textbf{\llmname{Gemma-2B} D2L hypernetwork.} The \llmname{Gemma-2B} D2L hypernetwork performs the same context-to-LoRA parameterization for the smaller \llmname{Gemma-2-2B-it} backbone~\citep{charakorn2026doc}.
    Its backbone-specific parameter generation allowed us to evaluate D2L consistently across model families and at a reduced parameter scale.
    We used the \nolinkurl{gemma_2b_d2l/checkpoint-20000} checkpoint, matching the identical training budget used for \llmname{Qwen}.
\end{itemize}
Since \method{} is a training-free approach, both the base models and the D2L hypernetworks remained frozen throughout evaluation, without requiring any additional parameter updates.

\subsection{Benchmark and Datasets}
\label{app:benchmark-datasets}

\paragraph{Benchmark.}
Table~\ref{tab:dataset-statistics} summarizes the detailed statistics of \benchmark{}, which comprises \textbf{2,944 contexts} and \textbf{12,932 QA pairs}.
Averaged across the five constituent datasets and weighted by context frequency, the mean length of the full interaction history is \textbf{3,811.9 tokens}, as measured by the \llmname{Qwen3-4B-Instruct-2507} tokenizer.
Per-dataset averages vary considerably, ranging from 328.3 tokens on SQuAD to 18,539.8 tokens on 2WikiMultihopQA, underscoring the substantial diversity in context length spanned by the benchmark.

\paragraph{Datasets.}
In total, we employed seven datasets spanning factual and reasoning-intensive settings.
The first five constitute \benchmark{}, while GSM8K and CRUXEval were additionally incorporated to examine whether \method{}, together with the compared baseline and oracle approaches, generalizes beyond factual QA to more challenging mathematical and code-reasoning tasks.
For all datasets, we constructed and annotated samples under our context-update setting, thereby ensuring that every method was evaluated under an identical protocol across domains.

\begin{figure}[t]
    \centering
    \includegraphics[width=\columnwidth]{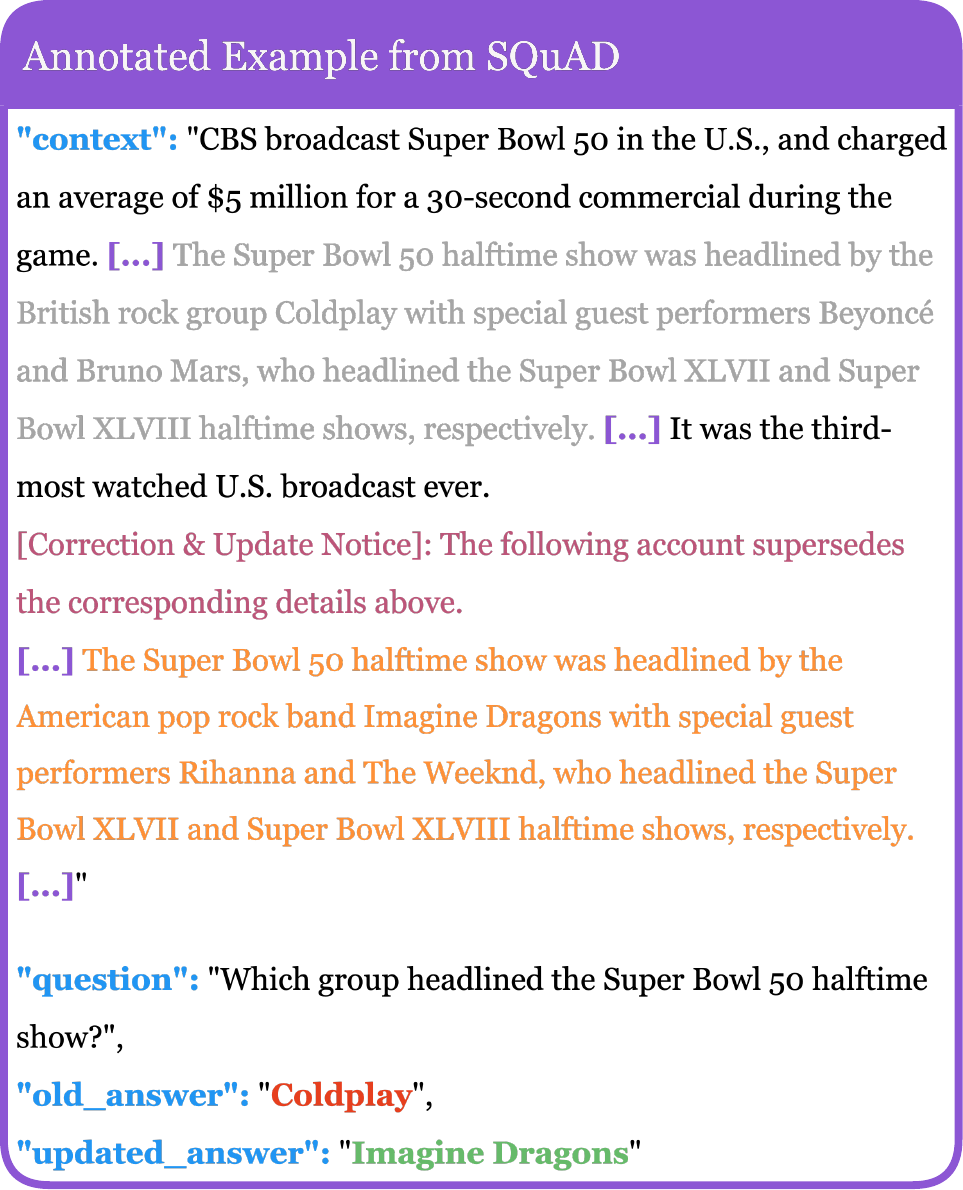}
    \caption{\textbf{Representative \benchmark{} example derived from SQuAD.} The example illustrates the original context, its subsequent update, and the associated question--answer annotations.}
    \label{fig:squad-example}
\end{figure}

\begin{figure}[t]
    \centering
    \includegraphics[width=\columnwidth]{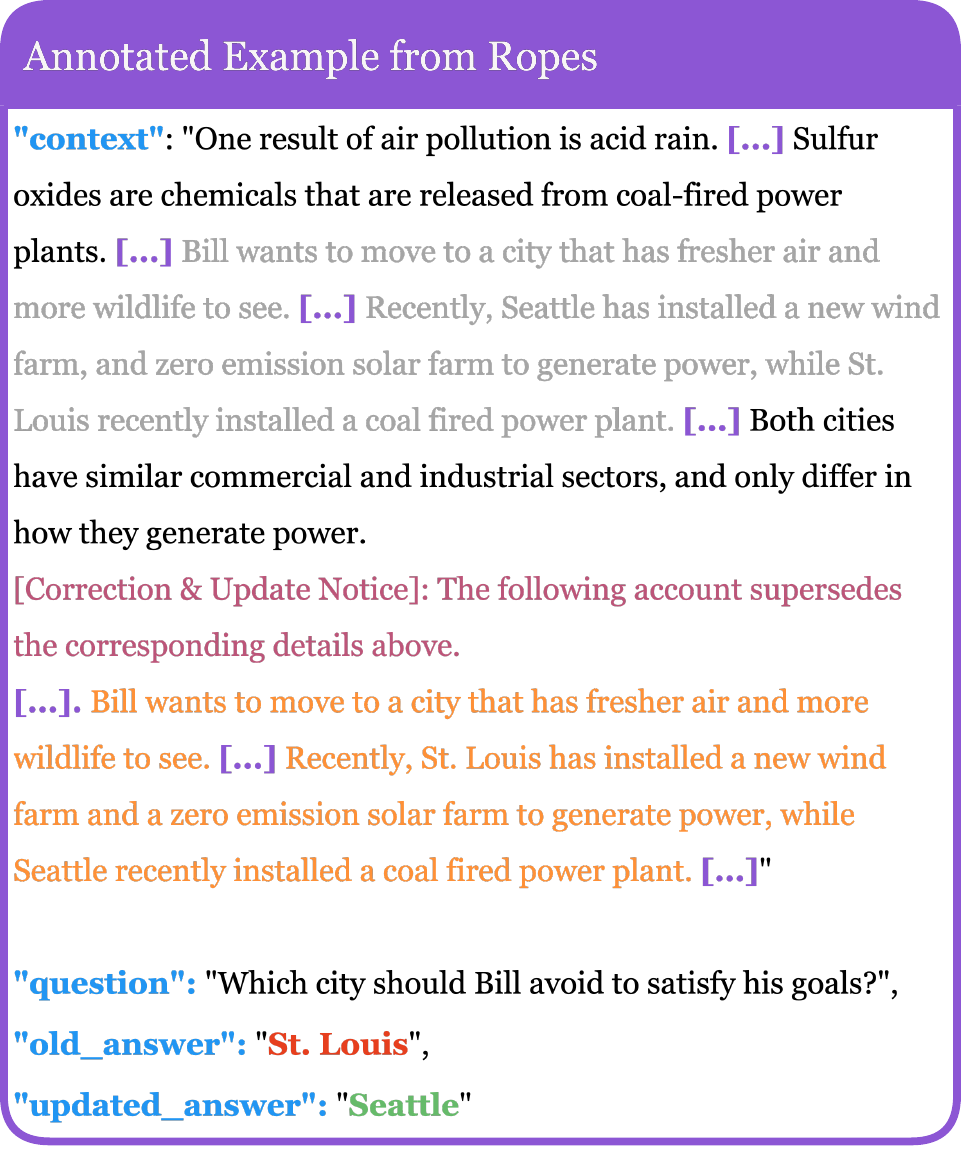}
    \caption{\textbf{Representative \benchmark{} example derived from ROPES.} The example illustrates a context update while preserving the causal reasoning structure of the original instance.}
    \label{fig:ropes-example}
\end{figure}

\begin{figure}[t]
    \centering
    \includegraphics[width=\columnwidth]{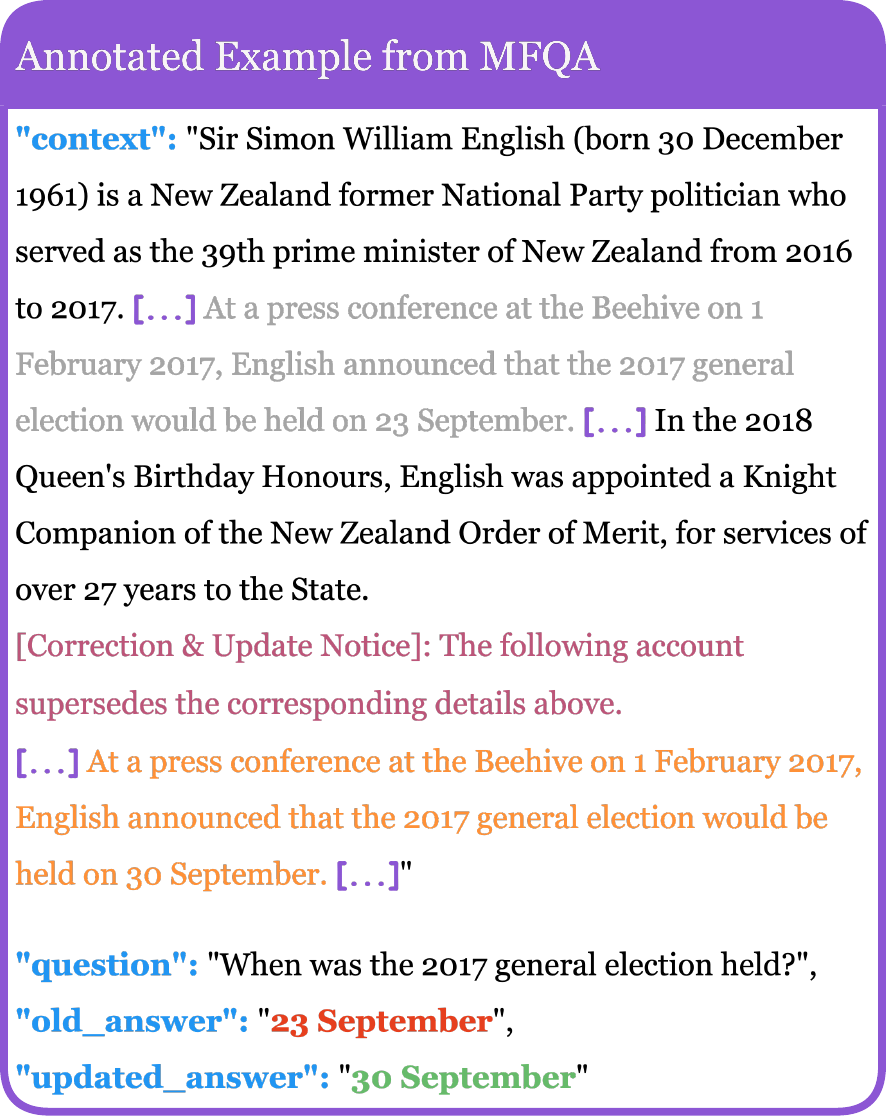}
    \caption{\textbf{Representative \benchmark{} example derived from MultiFieldQA-en.} The example illustrates an update within a long-form context while retaining its original document structure.}
    \label{fig:longbench-example}
\end{figure}

\begin{figure}[t]
    \centering
    \includegraphics[width=\columnwidth]{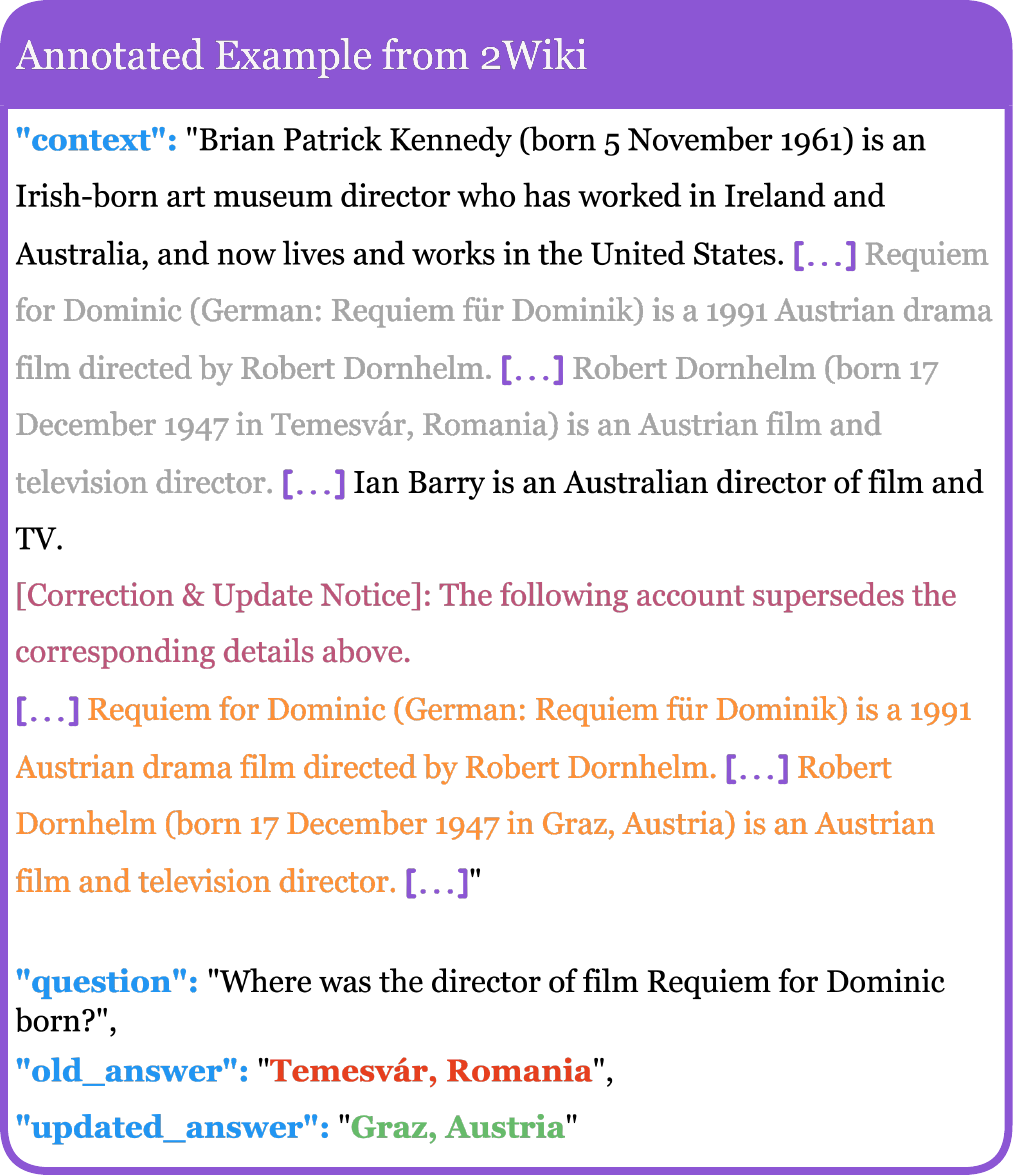}
    \caption{\textbf{Representative \benchmark{} example derived from 2WikiMultihopQA.} The example illustrates a context update with consistently revised multi-hop dependencies.}
    \label{fig:2wiki-example}
\end{figure}

\begin{figure}[t]
    \centering
    \includegraphics[width=\columnwidth]{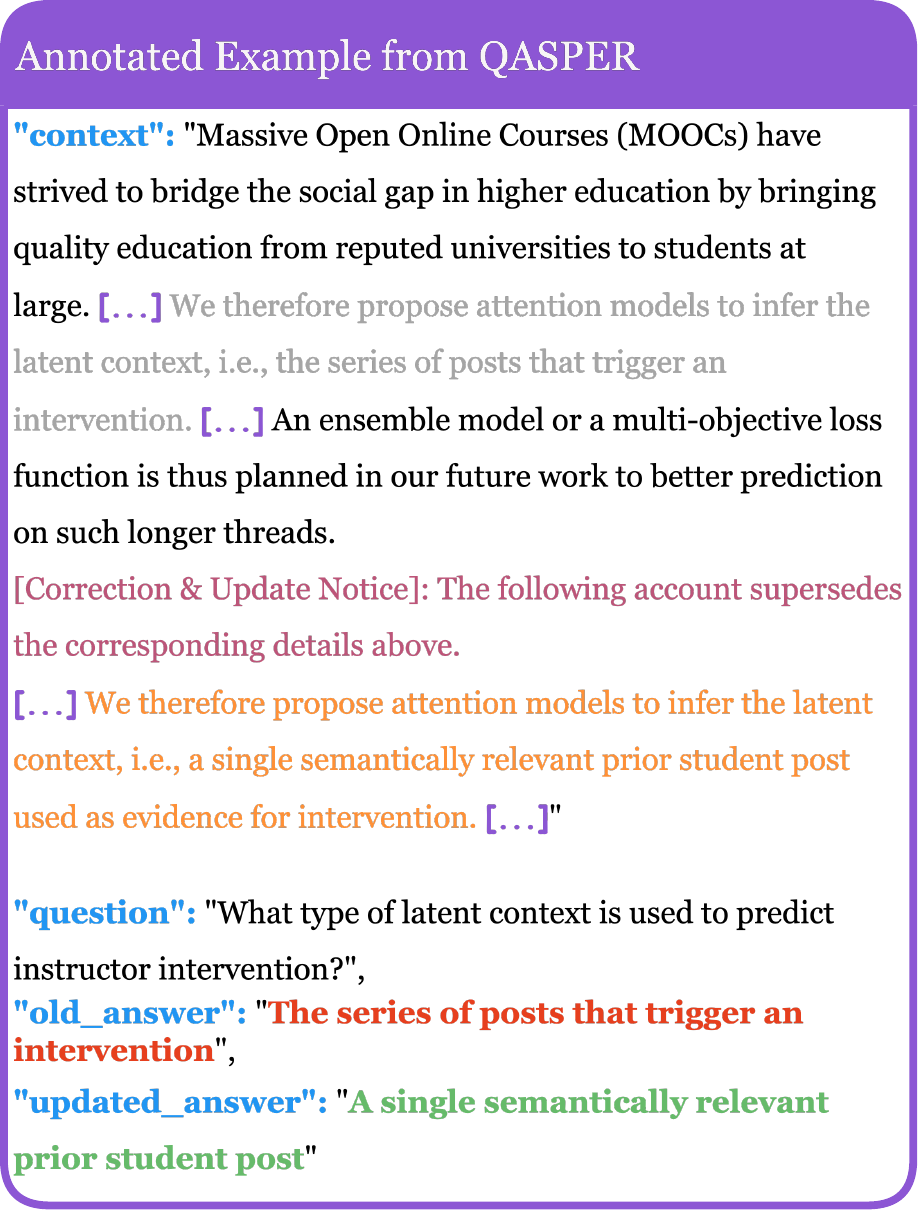}
    \caption{\textbf{Representative \benchmark{} example derived from QASPER.} The example illustrates an update to evidence distributed across a scientific document.}
    \label{fig:qasper-example}
\end{figure}

\begin{itemize}

\item \textbf{SQuAD.}
The Stanford Question Answering Dataset (SQuAD)~\citep{rajpurkar2016squad} is a widely used extractive question-answering benchmark comprising questions posed over Wikipedia passages, with answers grounded in corresponding textual spans.
Following our benchmark construction procedure, we annotated selected samples with context updates and their corresponding question.
We adopted the \texttt{rajpurkar/squad} dataset from Hugging Face as the underlying data source for this construction.
A representative annotated example is illustrated in Figure~\ref{fig:squad-example}.

\item \textbf{ROPES.}
Reasoning Over Paragraph Effects in Situations (ROPES)~\citep{lin2019reasoning} evaluates whether models can apply causal or qualitative relationships described in background passages to reason about novel, previously unseen situations.
We annotated selected samples by updating answer-relevant information and its associated dependencies, while preserving overall contextual consistency.
We adopted the \texttt{allenai/ropes} dataset from Hugging Face as the underlying data source.
A representative annotated example is illustrated in Figure~\ref{fig:ropes-example}.

\item \textbf{2WikiMultihopQA.}
2WikiMultihopQA~\citep{ho2020constructing} is a multi-hop question-answering dataset that requires compositional reasoning over multiple supporting facts drawn from interlinked Wikipedia passages.
We annotated selected samples by modifying answer-supporting facts and consistently propagating the corresponding updates throughout the associated reasoning dependencies.
We adopted the \texttt{xanhho/2WikiMultihopQA} dataset from Hugging Face as the underlying data source.
A representative annotated example is illustrated in Figure~\ref{fig:2wiki-example}.

\item \textbf{MultiFieldQA-en.}
MultiFieldQA-en is a single-document question-answering dataset from \rpsBenchmark{LongBench}~\citep{bai2024longbench}, containing long-form contexts collected from diverse domains and disciplines.
We annotated selected samples with context updates while preserving the remaining information and structure of the original long context.
We used the \texttt{multifieldqa\_en} subset of \texttt{THUDM\/LongBench} from Hugging Face.
A representative annotated example is shown in Figure~\ref{fig:longbench-example}.

\item \textbf{QASPER.}
QASPER~\citep{dasigi2021dataset} is a question-answering dataset built upon scientific papers, where answering a question may require synthesizing evidence distributed across multiple sections.
We annotated selected samples by updating answer-relevant evidence and its dependent information while preserving the overall structure and coherence of the original context.
We used the \texttt{allenai/qasper} dataset from Hugging Face.
A representative annotated example is shown in Figure~\ref{fig:qasper-example}.

\item \textbf{GSM8K.}
Grade School Math 8K (GSM8K)~\citep{cobbe2021training} contains linguistically diverse grade-school mathematical word problems that typically require multi-step arithmetic reasoning to solve correctly.
We annotated selected samples under our context-update setting to evaluate whether the same update mechanism generalized beyond question answering to mathematical reasoning.
We used the \texttt{openai/gsm8k} dataset from Hugging Face.

\item \textbf{CRUXEval.}
CRUXEval~\citep{gu2024cruxeval} is a code reasoning benchmark consisting of short Python functions paired with input--output examples, covering both input and output prediction tasks.
We annotated selected samples under our context-update setting to evaluate generalization to structured code reasoning.
We used the \texttt{cruxeval-org/cruxeval} dataset from Hugging Face.

\end{itemize}

Table~\ref{tab:dataset-statistics} reports the number of contexts, the number of question--answer pairs, and the average full-context length for the five datasets constituting \benchmark{}. Context length is measured using the \llmname{Qwen3-4B-Instruct-2507} tokenizer.

\begin{table}[tb]
\centering
\small
\begin{tblr}{width=\columnwidth,colspec={Q[l,wd=2.8cm]X[r]X[r]X[1.3,r]},cells={valign=m},colsep=2pt,rowsep=2pt}
\toprule
\SetRow{bg=black!6}
\textbf{Dataset} & \textbf{Contexts} & \textbf{QA pairs} & \textbf{Avg. context tokens} \\
\midrule
SQuAD             & 2,067 & 10,570 & 328.3 \\
ROPES             &   203 &  1,688 & 434.6 \\
2WikiMultihopQA   &   300 &    300 & 18,539.8 \\
MultiFieldQA-en   &   150 &    150 & 14,068.8 \\
QASPER            &   224 &    224 & 12,424.7 \\
\midrule
\benchmark{}        & 2,944 & 12,932 & 3,811.9 \\
\midrule
GSM8K             & 1,319 &  1,319 & 123.0 \\
CRUXEval          &   800 &    800 & 86.3 \\
\bottomrule
\end{tblr}
\caption{\textbf{Dataset statistics for \benchmark{} and the cross-task evaluation sets.} We report the numbers of contexts and question--answer pairs and the mean full-history length measured with the \llmname{Qwen3-4B-Instruct-2507} tokenizer.}
\label{tab:dataset-statistics}
\end{table}

\subsection{Baselines}
\label{app:baselines}

We compared \method{} with the following baselines:

\begin{itemize}

\item \textbf{Base Model w/o Context.}
The base model receives only the query and generates the response without access to any contextual information. This setting measures the model's parametric knowledge and serves as a lower-bound reference for context-dependent tasks.

\item \textbf{Base Model w/ Context (oracle).}
The base model receives the complete context history together with the query at inference time. Unlike parameterized methods, it directly accesses the textual context during generation and therefore serves as a full-context reference.

\item \textbf{D2L.}
Doc-to-LoRA (D2L)~\citep{charakorn2026doc} parameterizes textual context into LoRA adapters using a pretrained hypernetwork. We used the official pretrained hypernetwork checkpoint corresponding to each backbone. For contexts longer than the supported input length, we divided the full context history into chunks of at most 8,192 tokens and followed the original iterative layer-wise adapter generation procedure. The generated rank-8 LoRA adapters were applied to the MLP down-projection layer of each Transformer block.
The hypernetwork is trained across contexts with the same teacher--student
principle as context distillation: a context-conditioned teacher supplies the
target behavior, while the generated adapter enables the student to reproduce
that behavior without receiving the context at query time. Architecturally,
D2L aggregates variable-length context representations with a
Perceiver-style module and uses layer-specific output heads to generate the
LoRA factors. This amortizes the per-context optimization required by CD into
a single hypernetwork pass for a previously unseen context.

\item \textbf{Context Distillation (CD).}
Context Distillation (CD)~\citep{snell2022learning} transfers information from a context-conditioned teacher into model parameters through supervised distillation. For each context, we generated 20 auxiliary question--answer pairs in four rounds of five pairs and used them as distillation examples. We optimized a newly initialized rank-8 LoRA adapter with LoRA alpha 16 for 300 epochs using a learning rate of $10^{-4}$.

\item \textbf{Other oracle references.}
CD (oracle) uses the target evaluation query as its sole distillation query and the response generated by the full-context teacher as its target. A fresh query-specific rank-8 adapter is optimized for 300 iterations with the same LoRA alpha and learning rate as CD, and is then evaluated on that same query without textual context. D2L (oracle) parameterizes only the gold supporting context associated with the target query rather than the complete accumulated history. Both settings use query-specific information unavailable to the non-oracle methods and are therefore upper-bound references, not directly comparable deployment settings.

\item \textbf{AnyEdit.}
AnyEdit~\citep{jiang2025anyedit} is a knowledge-editing baseline that directly modifies model parameters to incorporate updated information. We used its AlphaEdit-ARE configuration. Edits were applied to the MLP down-projection layers of Transformer blocks 4--8. We divided the full context history into chunks of at most 8,192 tokens and used an edit-window size of 50 without overlap, 25 gradient steps per edit window, a learning rate of $0.5$, a weight decay of $0.001$, a clamp-norm factor of $4$, and an $L_2$ coefficient of $10$.

\end{itemize}

Except for Base Model w/ Context (oracle), all parameterized methods answered queries without receiving the full textual context again at inference time. Unless otherwise specified, all methods used the same base model and generation configuration.

\subsection{Evaluation}
\label{app:evaluation-setup}

Following Section~\ref{sec:benchmark-evaluation}, we evaluated answer coverage, semantic correctness, final-answer agreement, and locality. All answer-quality metrics were first computed for individual queries and then averaged within each dataset. When reporting an overall result across datasets, we used the \textbf{macro-average of the corresponding per-dataset scores} so that datasets with more question--answer pairs did not dominate the comparison.

\paragraph{ROUGE-L Overlap Metrics.}
For each query $i$, let $r_i=(r_{i,1},\ldots,r_{i,m_i})$ be the tokenized reference answer, and let $\hat{y}_i=(\hat{y}_{i,1},\ldots,\hat{y}_{i,n_i})$ be the tokenized model output. Denote the length of their longest common subsequence by
\begin{equation}
L_i=\operatorname{LCS}(r_i,\hat{y}_i).
\label{eq:rougel-lcs}
\end{equation}
ROUGE-L Recall measures how much of the reference answer is covered by the model output, whereas ROUGE-L Precision measures how much of the model output is supported by the reference answer:
\begin{equation}
R_i^{\mathrm{L}}=\frac{L_i}{m_i},
\qquad
P_i^{\mathrm{L}}=\frac{L_i}{n_i}.
\label{eq:rougel-recall-precision}
\end{equation}
Their harmonic mean gives ROUGE-L F1:
\begin{equation}
F_{1,i}^{\mathrm{L}}
=
\frac{2P_i^{\mathrm{L}}R_i^{\mathrm{L}}}
{P_i^{\mathrm{L}}+R_i^{\mathrm{L}}},
\label{eq:rougel-f1}
\end{equation}
where the score is defined as zero when the denominator is zero. These overlap metrics are computed with the \texttt{rouge-score} implementation and stemming enabled. Because the longest common subsequence preserves token order without requiring matched tokens to be contiguous, these metrics accommodate short intervening phrases while still rewarding structurally consistent answer overlap. Among the three, ROUGE-L Recall serves as our primary lexical metric, as reference answers are generally concise and answer coverage is more important than matching the reference length or wording exactly. This choice is further motivated by a systematic mismatch between instruction-tuned models and span-style references: models such as \llmname{Qwen} and \llmname{Gemma} typically produce complete, fluent sentences rather than terse answer spans, so even semantically correct responses tend to be substantially longer than the reference. Under this mismatch, Precision and F1 disproportionately penalize verbose yet correct generations, making them less reliable as standalone indicators of answer quality. We nonetheless retain Precision and F1 as complementary diagnostics: Precision additionally penalizes unrelated or unnecessarily long generations, while F1 summarizes the balance between coverage and conciseness. Together, these variants allow us to distinguish omission of reference content from the inclusion of extraneous material. For a dataset $d$ containing query set $\mathcal{Q}_d$, each metric $M\in\{R^{\mathrm{L}},P^{\mathrm{L}},F_1^{\mathrm{L}}\}$ is aggregated as
\begin{equation}
M_d=\frac{1}{|\mathcal{Q}_d|}\sum_{i\in\mathcal{Q}_d}M_i.
\label{eq:rougel-dataset-aggregation}
\end{equation}

\paragraph{Locality.}
Locality measures whether incorporating an update damages information that should remain unchanged. For dataset $d$, let $\mathcal{U}_d\subseteq\mathcal{Q}_d$ denote the queries whose correct answers are unaffected by the update. Because their original and current reference answers are identical, any reduction in answer coverage after context parameterization indicates interference with unaffected information. We computed Locality as the mean ROUGE-L Recall on this unaffected-query subset:
\begin{equation}
\operatorname{Locality}_d
=
\frac{1}{|\mathcal{U}_d|}
\sum_{i\in\mathcal{U}_d}
R_i^{\mathrm{L}}.
\label{eq:locality}
\end{equation}
Update-affected queries are excluded from $\mathcal{U}_d$. Our locality evaluation uses 50 annotated SQuAD contexts containing 754 queries in total. Exactly one answer is changed in each context, leaving 704 unaffected queries for computing Eq.~\ref{eq:locality}. Thus, a higher Locality score means that the method more faithfully preserves answers supported by facts that were not modified, while a lower score indicates collateral interference introduced by context parameterization, parameter editing, or decoding. Locality is reported on a $0$--$100$ scale in the figures.

For D2 Semantic Equivalence and D3 Final-Answer Agreement, we used \llmname{GPT-5.6-Sol} under unified evaluation prompts and criteria. For efficiency, we separately measured latency and peak GPU memory during context parameterization and query generation, as detailed in Appendix~\ref{app:efficiency-measurement}.

\paragraph{Evaluation Rubric.}
\label{app:evaluation-rubric}

We used \llmname{GPT-5.6-Sol} to evaluate each generated answer under two independent rubrics corresponding to D2 Semantic Equivalence and D3 Final-Answer Agreement. Each evaluation instance contained the query, the current reference answer determined by the full context history, and the model output. The judge did not receive $C_{\mathrm{old}}$, $C_{\mathrm{update}}$, or $C_{\mathrm{full}}$. We used a temperature of 0, allowed at most 300 output tokens, and required a JSON object. Each rubric was evaluated in a separate call. The complete evaluation prompts are provided in Appendix~\ref{app:prompts}.

\paragraph{Judgment Output and Aggregation.}
\label{app:rubric-aggregation}

The two rubrics were evaluated independently, each targeting a distinct facet of answer quality.
For each rubric, the judge returned a binary \texttt{yes}/\texttt{no} decision, a brief rationale justifying the verdict.
A query received an overall LLM-as-a-Judge score of 1 \textbf{only when both D2 Semantic Equivalence and D3 Final-Answer Agreement were simultaneously satisfied}; otherwise, it received a score of 0.
We report the mean binary score across queries within each dataset.

\subsection{Implementation Details}
\label{app:implementation-details}

All experiments were conducted on two NVIDIA A800 GPUs using bfloat16 precision.

\subsubsection{Exact LoRA Composition and Rank}
\label{app:lora-composition}

As defined in Section~\ref{sec:background}, each context adapter represents an effective update
$\Delta \mathrm{W}_C=B_CA_C$, where
$A_C\in\mathbb{R}^{r_C\times d_{\mathrm{in}}}$ and
$B_C\in\mathbb{R}^{d_{\mathrm{out}}\times r_C}$.
Writing $c_{\mathrm{full}}=\alpha+\beta$ and
$c_{\mathrm{old}}=-\beta$, Eq.~\ref{eq:global_adapter} becomes
\begin{equation}
\Delta \mathrm{W}_g
=c_{\mathrm{full}}B_{\mathrm{full}}A_{\mathrm{full}}
+c_{\mathrm{old}}B_{\mathrm{old}}A_{\mathrm{old}}.
\label{eq:global-effective-update}
\end{equation}
We represent this sum exactly with the concatenated factors
\begin{equation}
\begin{aligned}
A_g & =
\begin{bmatrix}
A_{\mathrm{full}}\\
A_{\mathrm{old}}
\end{bmatrix},\\
B_g & =
\begin{bmatrix}
c_{\mathrm{full}}B_{\mathrm{full}} &
c_{\mathrm{old}}B_{\mathrm{old}}
\end{bmatrix}.
\end{aligned}
\label{eq:global-factor-concatenation}
\end{equation}
Consequently, $B_gA_g$ is exactly Eq.~\ref{eq:global-effective-update},
without element-wise subtraction of LoRA factors, dense-weight merging, or
recompression to rank 8. The stored factorization width is therefore
$r_g=r_{\mathrm{full}}+r_{\mathrm{old}}$; when both inputs are single
rank-8 adapters, $r_g=16$. The algebraic matrix rank satisfies
$\operatorname{rank}(\Delta \mathrm{W}_g)\leq r_g$ and may be lower, so
$r_g$ refers to the implemented LoRA factorization width rather than a claim
that the effective matrix necessarily has full rank.

For a context $C$ divided into $n_C$ chunks, D2L first generates one
rank-$r_0$ adapter per chunk ($r_0=8$ in our experiments) and aggregates the
chunk updates by the same rank-wise concatenation. Thus,
$r_C=n_Cr_0$ and the global factorization width is
$r_g=(n_{\mathrm{full}}+n_{\mathrm{old}})r_0$. If the D2L checkpoint enables
its learned rank-$r_0$ LoRA bias block, that block is appended once to each
context adapter, giving $r_C=(n_C+1)r_0$ and
$r_g=(n_{\mathrm{full}}+n_{\mathrm{old}}+2)r_0$ instead. These concatenated
factors are applied directly during the LoRA forward pass; the corresponding
dense matrices are never materialized. Accordingly, the adapter storage and
LoRA multiplication cost scale linearly with the implemented factorization
width $r_g$.

\section{Hyperparameters}
\label{app:hyperparameters}

We separated the backbone and checkpoint configuration, the context-parameterization execution mode, and the \method{}-specific inference hyperparameters. We reported inference settings only; the training configuration of the released D2L hypernetworks was not included. All models and hypernetworks remained frozen and were loaded in bfloat16 precision. Table~\ref{tab:model-d2l-settings} summarizes the backbone and D2L checkpoint configurations.

\begin{table}[tb]
\centering
\small
\begin{tblr}{width=\columnwidth,colspec={X[1.4,l]X[1.65,l]X[0.65,l]X[l]},cells={valign=m},colsep=2pt,rowsep=2pt}
\toprule
\SetRow{bg=black!6}
\textbf{Base model} & \textbf{D2L checkpoint} & \textbf{Base rank} & \textbf{Target} \\
\midrule
\llmname{Qwen3-4B} & \texttt{Qwen D2L-20k} & 8 & MLP down \\
\llmname{Gemma-2-2B} & \texttt{Gemma D2L-20k} & 8 & MLP down \\
\bottomrule
\end{tblr}
\caption{\textbf{Backbone and D2L checkpoint configurations.} All models use bfloat16 precision, and the generated rank-8 LoRA adapters are applied to the MLP down-projection layers.}
\label{tab:model-d2l-settings}
\end{table}

\paragraph{Context parameterization.}
For contexts exceeding the supported input length, we preserved the complete context history by dividing it into near-equal contiguous chunks of at most 8,192 tokens. We used the Iterative D2L inference mode for all reported experiments. Table~\ref{tab:d2l-inference-modes} also describes the available Batched mode for completeness; no experimental metrics are reported for that mode. In \textbf{Iterative} mode, the official layer-wise procedure processed the Transformer-layer representations sequentially. In \textbf{Batched} mode, the layer dimension was folded into the batch dimension and the corresponding representations were processed jointly. This execution-mode choice was independent of context chunking: both modes support the same complete-context chunks and generate rank-8 LoRA adapters with the same structure. Here, 8 denotes the base rank of each chunk adapter, not the width of the final composed adapter. Chunk aggregation and global update composition increase the factorization width as detailed in Appendix~\ref{app:lora-composition}. For \method{}, \textbf{the query was used only to activate memory evidence and was never passed to the D2L hypernetwork}.

\begin{table}[tb]
\centering
\small
\begin{tblr}{width=\columnwidth,colspec={X[1.2,l]X[c]X[c]},cells={valign=m},colsep=2pt,rowsep=2pt}
\toprule
\SetRow{bg=black!6}
\textbf{Setting} & \textbf{Iterative} & \textbf{Batched} \\
\midrule
Context coverage & Complete history & Complete history \\
Maximum chunk length & 8,192 & 8,192 \\
Chunk construction & Near-equal, contiguous & Near-equal, contiguous \\
Layer execution & Sequential & Joint batched pass \\
Base LoRA rank & 8 & 8 \\
Target module & \texttt{down\_proj} & \texttt{down\_proj} \\
Hypernetwork input & Context/evidence only & Context/evidence only \\
\bottomrule
\end{tblr}
\caption{\textbf{D2L inference modes for context parameterization.} Iterative and batched execution differ only in how Transformer-layer representations are processed; both support complete-context chunking and the same base-rank configuration. Only Iterative results are reported.}
\label{tab:d2l-inference-modes}
\end{table}

\paragraph{\method{} and generation settings.}
Table~\ref{tab:plume-implementation-settings} summarizes the method-specific and generation hyperparameters.
\method{} separately parameterizes $C_{\mathrm{old}}$ and $C_{\mathrm{full}}$ and constructs the global update LoRA with $\alpha=1$ and $\beta=0.75$ in Eq.~\ref{eq:global_adapter}. Memory units are formed independently of the 8,192-token full-context chunks while preserving sentence and paragraph boundaries whenever possible. \method{} activates one memory unit using the relevance score defined below and a recency margin of $\delta=0.5$. We used $\lambda_{\max}=1$ and $\tau=0.3$ for adaptive memory evidence decoding, greedy decoding with a temperature of 0, and at most 256 new tokens for both backbones and all comparison methods.

For the component ablations in Figure~\ref{fig:ablation}, the \textbf{memory-only} variant removed the global update representation while retaining the same memory segmentation, query-dependent evidence activation, and hypernetwork parameterization $\Delta \mathrm{W}_e(q)=H_\phi(m^\star(q))$; it generated predictions using the resulting evidence adapter alone. The \textbf{global-only} variant retained $\Delta \mathrm{W}_g$ while removing evidence activation, evidence parameterization, and adaptive fusion. In both variants, as in the complete method, no memory-evidence text was provided to the target language model.

For lexical routing, text is lowercased and tokenized with the regular expression \texttt{[a-z0-9]+}. Let $Q$ be the set of query terms, $c_{i,t}$ the count of term $t$ in memory unit $m_i$, and $w_t$ its document-level inverse-frequency weight. With $k_1=1.5$ and $\mathrm{W}=\sum_{t\in Q}w_t$, the lexical score is
\begin{equation}
\begin{aligned}
s_{\mathrm{lex}}(q,m_i)
&=\frac{1}{2}\frac{\sum_{t:c_{i,t}>0}w_t}{\mathrm{W}}
  +\frac{1}{2}\frac{F_i}{F_i+\mathrm{W}}, \\
F_i
&=\sum_{t:c_{i,t}>0}w_t
  \frac{c_{i,t}(k_1+1)}{c_{i,t}+k_1}.
\end{aligned}
\label{eq:lexical-router}
\end{equation}
Let $s_{\max}=\max_i s_{\mathrm{lex}}(q,m_i)$. We form the candidate set
\begin{equation}
\mathcal{I}_{\delta}
=\left\{i\,\middle|\,
s_{\mathrm{lex}}(q,m_i)\geq s_{\max}-\delta
\right\}.
\label{eq:recency-margin}
\end{equation}
We use $\delta=0.5$ and activate the temporally latest unit in $\mathcal{I}_{\delta}$. Temporal order follows each unit's position in $C_{\mathrm{full}}$. Thus, recency takes precedence only among units whose relevance scores are within $\delta$ of the maximum. If the query-term set is empty, all units receive a score of zero and the temporally latest unit is activated.

\begin{table}[t]
\centering
\small
\begin{tblr}{width=\columnwidth,colspec={X[2,l]X[c]},cells={valign=m},colsep=2pt,rowsep=2pt}
\toprule
\SetRow{bg=black!6}
\textbf{Setting} & \method{} \\
\midrule
Maximum context-chunk length & 8,192 \\
Evidence activation & Relevance with recency margin \\
Number of activated units & 1 \\
Recency margin $\delta$ & 0.5 \\
$\alpha$ / $\beta$ & 1 / 0.75 \\
$\lambda_{\max}$ / $\tau$ & 1 / 0.3 \\
Maximum new tokens & 256 \\
Temperature & 0 \\
Decoding & Greedy \\
\bottomrule
\end{tblr}
\caption{\textbf{\method{}-specific and generation hyperparameters.} Memory units are constructed from context structure, so their number varies with the length and organization of each context.}
\label{tab:plume-implementation-settings}
\end{table}

\section{More Experiments}
\label{app:more-experiments}

This section presents supplementary discussion of the main results together
with analyses of \method{}'s hyperparameter sensitivity and computational
efficiency.

\subsection{Additional Analysis of the Main Results}
\label{app:additional-result-analysis}

\begin{figure}[t]
\centering
\includegraphics[width=\columnwidth]{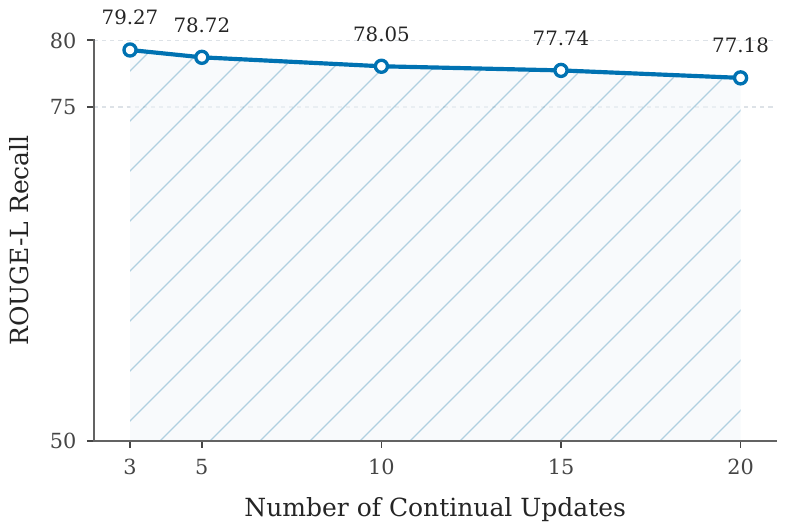}
\caption{\textbf{Performance under continual updates.} ROUGE-L Recall decreases gradually as the number of updates increases from 3 to 20.}
\label{fig:continual-updates}
\end{figure}

\paragraph{Continual-update performance.}
\label{app:multi-update-experiments}
We further evaluated \method{} as the number of continual updates increased. Figure~\ref{fig:continual-updates} reports ROUGE-L Recall of 79.27, 78.72, 78.05, 77.74, and 77.18 after 3, 5, 10, 15, and 20 updates, respectively. The gradual decline of 2.09 points from 3 to 20 updates suggests that \method{} remains effective as successive revisions accumulate, although longer update histories still introduce additional difficulty. These results extend the single-transition evaluation in the main benchmark and support the applicability of the method to continual context evolution.

\paragraph{Cross-task generalization.}
GSM8K and CRUXEval differ from the five \benchmark{} QA datasets in both output
structure and reasoning process: the former requires multi-step arithmetic,
whereas the latter requires deterministic program reasoning. \method{}'s gains on
both tasks and both backbones show that its benefit is not restricted to a
particular extractive or document-QA format, but extends to broader continual
context-update scenarios.

\paragraph{Overall comparison and component roles.}
The overall comparison jointly considers ROUGE-L Recall, LLM-as-a-Judge,
update/generation latency and memory, and Locality. \method{}'s advantage is
distributed across these dimensions rather than being driven by a single
answer-quality metric. The component analysis further clarifies the division
of labor: the global update representation preserves the full context state
while emphasizing the latest change, and query-activated memory evidence
provides a local parameter view that reduces interference from information
irrelevant to the current query. Their combination accounts for the stronger
balance between update adoption and preservation of unaffected knowledge.

\paragraph{Radar-chart normalization.}
For Figure~\ref{fig:overall_radar}, let $\mathcal{M}$ contain the four plotted methods: \method{}, CD, CD (oracle), and AnyEdit. Let $x_{m,j}$ denote method $m$'s value on axis $j$, with dataset-specific axes normalized separately. Let $\mathcal{J}_{+}$ contain the answer-quality and Locality metrics and $\mathcal{J}_{-}$ contain the latency and memory metrics. We first orient each metric so that larger values are better:
\begin{equation}
u_{m,j}=\begin{cases}
x_{m,j}, & j\in\mathcal{J}_{+},\\
x_{m,j}^{-1}, & j\in\mathcal{J}_{-}.
\end{cases}
\end{equation}
The plotted value is then
\begin{equation}
z_{m,j}=\frac{u_{m,j}}{\max_{k\in\mathcal{M}}u_{k,j}}.
\end{equation}
Thus, the best of the four methods receives 1 on each axis. For the positive latency and memory measurements, this is equivalent to dividing the smallest value by the method's value.

\paragraph{Ablation normalization.}
Figure~\ref{fig:ablation} uses a different reference: each score is divided by the corresponding Base Model w/ Context (oracle) score on the same dataset and backbone, then multiplied by 100. For \method{}, this gives SQuAD Recall $100\times80.65/95.86=84.13$, 2Wiki Recall $100\times47.24/60.11=78.59$, and SQuAD Locality $100\times84.24/96.42=87.37$. The reported average normalized ROUGE-L Recall of 81.36 is the mean of the two normalized Recall scores; the same averaging rule yields 57.28 and 63.95 for the memory-only and global-only variants, respectively.

\subsection{Candidate-Rank Analysis}
\label{app:topk-analysis}

\begin{figure}[t]
\centering
\includegraphics[width=\columnwidth]{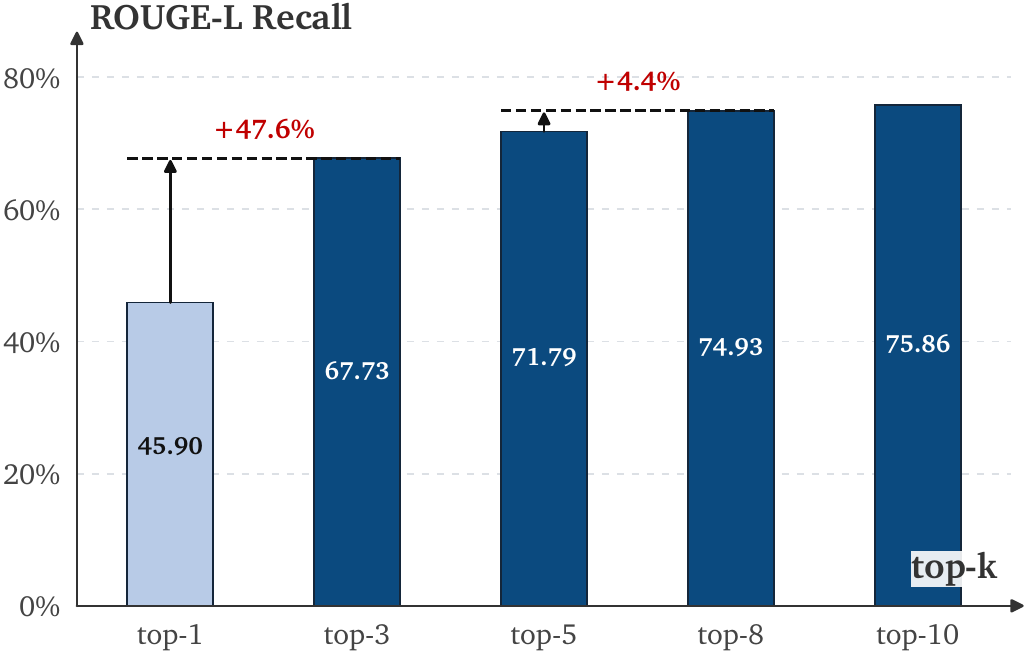}
\caption{\textbf{Teacher-forced candidate-rank analysis.} Token-level ROUGE-L Recall increases sharply from top-1 to top-5 and then begins to saturate through top-10.}
\label{fig:topk-analysis}
\end{figure}

To determine whether continual-update interference erases valid information or merely suppresses it, we extended the teacher-forced token analysis in Section~\ref{sec:introduction} from top-1 to top-10 candidates. Figure~\ref{fig:topk-analysis} shows that ROUGE-L Recall rises from \textbf{45.90\%} at top-1 to \textbf{67.73\%} at top-3 and \textbf{71.79\%} at top-5, then increases more gradually to 74.93\% and 75.86\% at top-8 and top-10. The large early gains, followed by saturation, indicate that \textbf{much of the currently valid answer remains near the top of the output distribution but is outranked by interfering candidates}. This result supports targeted reweighting rather than treating the missing top-1 prediction as complete information loss.

\subsection{Sensitivity to \texorpdfstring{$\beta$}{β}}
\label{app:beta-sensitivity}

\begin{figure}[t]
\centering
\includegraphics[width=\columnwidth]{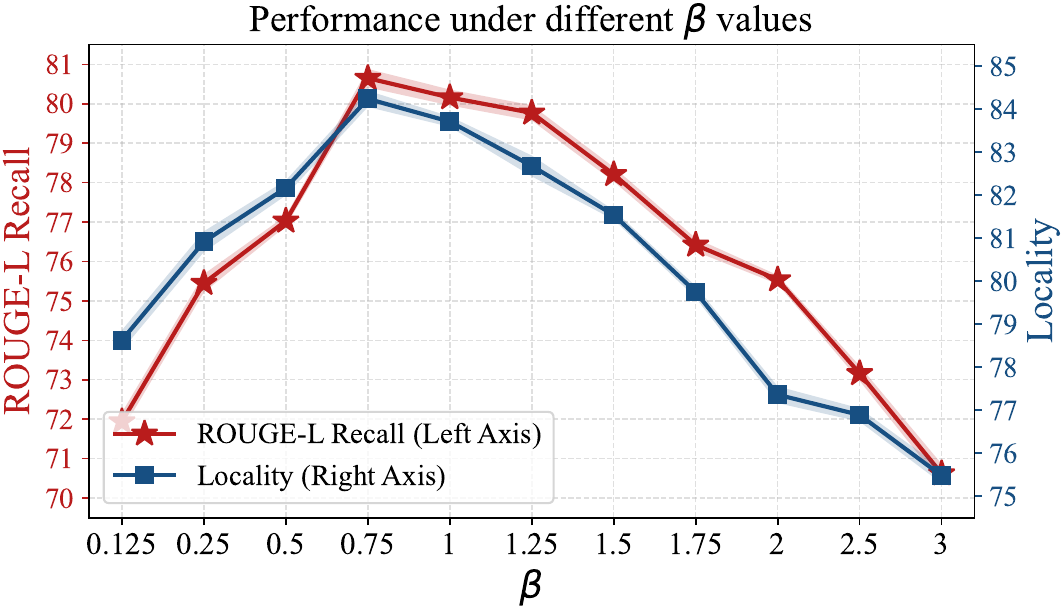}
\caption{\textbf{Sensitivity to the global update coefficient $\beta$.} Recall and Locality achieve their best joint result at $\beta=0.75$ and decrease when the update direction is under- or over-amplified.}
\label{fig:beta-sensitivity}
\end{figure}

The coefficient $\beta$ controls how strongly the latest-update direction is amplified within the global representation. As shown in Figure~\ref{fig:beta-sensitivity}, increasing $\beta$ from 0.125 to 0.75 raises ROUGE-L Recall from 71.94 to \textbf{80.65} and Locality from 78.63 to \textbf{84.24}. Both metrics then decline as amplification becomes excessive, reaching 70.62 and 75.48 at $\beta=3$. We therefore use $\beta=0.75$. Importantly, this optimum is obtained by strengthening the update direction inside the full-context representation, rather than replacing that representation with query-local evidence. The result supports the intended division of labor: global context remains the primary state, while local evidence only corrects query-specific suppression.

\subsection{Sensitivity to \texorpdfstring{$\tau$}{τ}}
\label{app:tau-sensitivity}

\begin{figure}[t]
\centering
\includegraphics[width=\columnwidth]{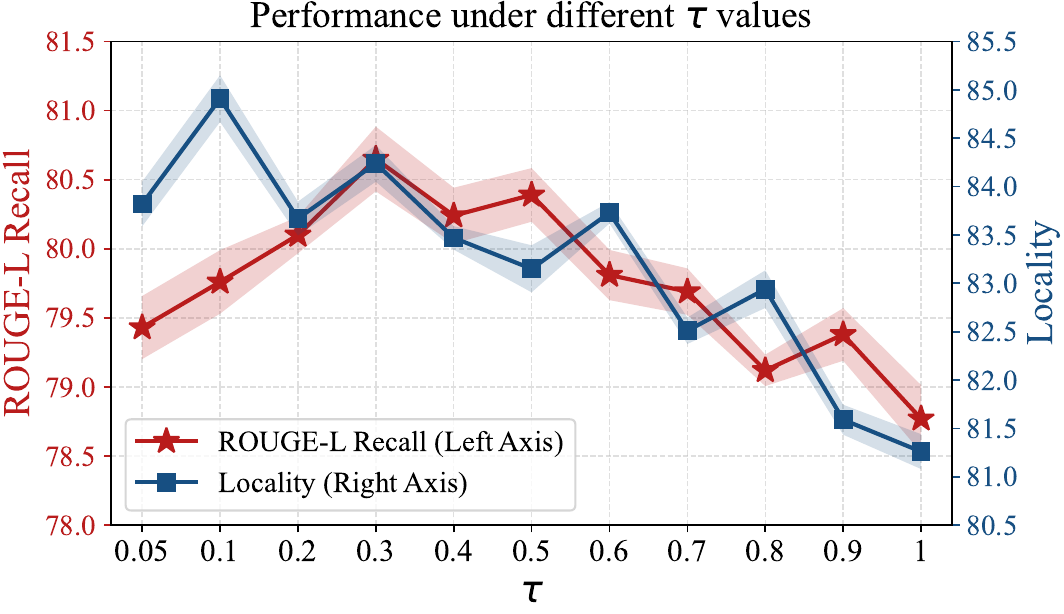}
\caption{\textbf{Sensitivity to the adaptive-gating parameter $\tau$.} The selected value $\tau=0.3$ maximizes Recall while maintaining strong Locality.}
\label{fig:tau-sensitivity}
\end{figure}

The parameter $\tau$ determines how quickly the adaptive evidence weight grows with predictive divergence. Figure~\ref{fig:tau-sensitivity} shows that ROUGE-L Recall reaches its maximum of \textbf{80.65} at $\tau=0.3$, while Locality reaches 84.24. The neighboring settings $\tau=0.2$ and $\tau=0.4$ yield lower Recall values of 80.10 and 80.24, respectively; beyond this region, performance generally weakens, with Recall/Locality falling to 78.77/81.26 at $\tau=1$. We therefore select $\tau=0.3$ as the best operating point for answer recovery while retaining strong unaffected-information preservation.

\subsection{Sensitivity to \texorpdfstring{$\lambda_{\max}$}{λmax}}
\label{app:lambda-sensitivity}

\begin{figure}[t]
\centering
\includegraphics[width=\columnwidth]{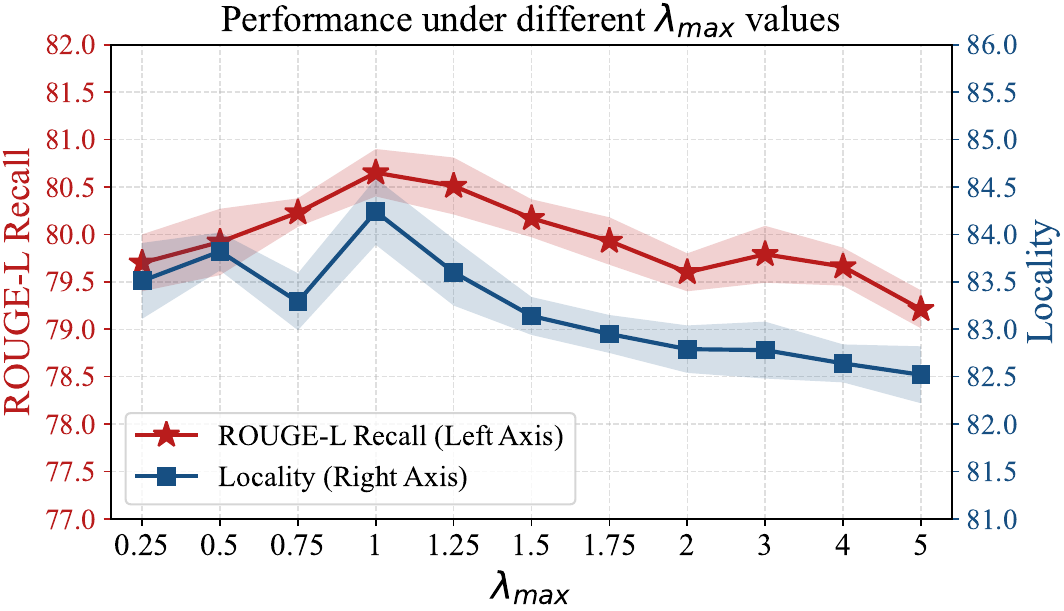}
\caption{\textbf{Sensitivity to the maximum evidence weight $\lambda_{\max}$.} Recall and Locality achieve their strongest joint balance at $\lambda_{\max}=1$ and decline when local evidence is assigned excessive influence.}
\label{fig:lambda-sensitivity}
\end{figure}

The cap $\lambda_{\max}$ directly limits the maximum contribution of query-relevant memory evidence. Figure~\ref{fig:lambda-sensitivity} shows the strongest joint result at $\lambda_{\max}=1$, where Recall and Locality reach \textbf{80.65} and \textbf{84.24}, respectively. Recall remains near 80 over a moderate neighborhood, but increasing the cap to 5 reduces Recall/Locality to 79.21/82.52. Thus, allowing local evidence to dominate does not improve the model; it instead erodes the preservation supplied by the global context state. Together with the $\beta$ analysis, this directly addresses the concern that a query-dependent method might collapse into local-evidence decoding: the empirically preferred configuration strengthens the full-context update direction and explicitly caps local evidence at a complementary level.

\begin{table*}[!t]
\centering
\small
\providecolor{effheaderbg}{HTML}{F9F6FF}
\begin{tblr}{
    width=\textwidth,
    colspec={Q[c,wd=3.4cm] *{4}{X[c]}},
    rowsep=2pt,
    colsep=5pt,
    row{1-2}={bg=effheaderbg, halign=c, valign=m},
    row{1}={font=\bfseries},
    cell{3}{2-5}={fg=gray},
    hline{1,3}={wd=1pt,solid},
    hline{2}={2-3}{wd=0.5pt,solid,leftpos=0,rightpos=-0.5,endpos},
    hline{2}={4-5}{wd=0.5pt,solid,leftpos=-0.5,rightpos=0,endpos},
    hline{4,7}={wd=0.5pt,solid},
    hline{8}={wd=1pt,solid},
}
\SetCell[r=2]{halign=c,valign=m,font=\bfseries} Method
& \SetCell[c=2]{halign=c,valign=m} SQuAD & {}
& \SetCell[c=2]{halign=c,valign=m} 2Wiki & {} \\
& Mem. (GiB) $\downarrow$
& Lat. (s/document) $\downarrow$
& Mem. (GiB) $\downarrow$
& Lat. (s/document) $\downarrow$ \\
CD (oracle)      & 3.568 & 56.600 & 22.376 & 950.085 \\
AnyEdit          & \underline{6.332} & 18.754 & 55.028 & 4099.473 \\
CD               & 48.708 & 451.827 & \underline{51.956} & 489.059 \\
D2L              & \textbf{0.438} & \underline{0.466} & \textbf{11.536} & \underline{1.841} \\
\method{}   & \textbf{0.438} & \textbf{0.447} & \textbf{11.536} & \textbf{1.832} \\
\end{tblr}
\caption{
\textbf{Update efficiency on SQuAD and 2Wiki.}
Memory is the peak additional CUDA memory allocated during context
parameterization, and latency is the mean update time per document. The two
base-model references are omitted because they incur no update-stage cost.
Oracle methods are excluded from ranking; bold and underlined values denote
the best and second-best non-oracle results, respectively. Lower values are
better, and tied best values are both bolded.
}
\label{tab:update-efficiency}
\end{table*}

\begin{table*}[!t]
\centering
\small
\providecolor{effheaderbg}{HTML}{F9F6FF}
\begin{tblr}{
    width=\textwidth,
    colspec={Q[c,wd=3.4cm] *{4}{X[c]}},
    rowsep=2pt,
    colsep=5pt,
    row{1-2}={bg=effheaderbg, halign=c, valign=m},
    row{1}={font=\bfseries},
    cell{3-4}{2-5}={fg=gray},
    hline{1,3}={wd=1pt,solid},
    hline{2}={2-3}{wd=0.5pt,solid,leftpos=0,rightpos=-0.5,endpos},
    hline{2}={4-5}{wd=0.5pt,solid,leftpos=-0.5,rightpos=0,endpos},
    hline{5,9}={wd=0.5pt,solid},
    hline{10}={wd=1pt,solid},
}
\SetCell[r=2]{halign=c,valign=m,font=\bfseries} Method
& \SetCell[c=2]{halign=c,valign=m} SQuAD & {}
& \SetCell[c=2]{halign=c,valign=m} 2Wiki & {} \\
& Mem. (GiB) $\downarrow$
& Lat. (s/query) $\downarrow$
& Mem. (GiB) $\downarrow$
& Lat. (s/query) $\downarrow$ \\
w/ Context (oracle)  & 0.208 & 1.972 & 1.749 & 1.194 \\
CD (oracle)          & 0.057 & 1.663 & 0.053 & 4.709 \\
w/o Context      & 0.047 & 6.467 & \underline{0.042} & \textbf{0.924} \\
AnyEdit          & 0.042 & 4.789 & 0.046 & \underline{4.709} \\
CD               & \textbf{0.026} & 1.906 & \textbf{0.032} & 5.718 \\
D2L              & \underline{0.039} & \textbf{1.577} & 0.045 & 5.079 \\
\method{}   & 0.059 & \underline{1.592} & 0.068 & 5.152 \\
\end{tblr}
\caption{
\textbf{Generation efficiency on SQuAD and 2Wiki.}
Memory is the peak additional CUDA memory allocated during generation, and
latency is the mean generation time per query. Oracle methods are excluded
from ranking; bold and underlined values denote the best and second-best
non-oracle results, respectively. Lower values are better.
}
\label{tab:generation-efficiency}
\end{table*}

\subsection{Efficiency Measurement}
\label{app:efficiency-measurement}

We followed the efficiency protocol of D2L~\citep{charakorn2026doc} and separately measured context update and query generation. \textbf{Update Latency} is the mean wall-clock time required to transform one full context history into the reusable parameterized state of a method. It includes all method-specific context processing and parameter construction performed before query answering. Sequential components of the same update were summed for each context before averaging across contexts. \textbf{Generation Latency} is the mean wall-clock time required to generate an answer after the update stage is complete. The total generation time was divided by the number of evaluated queries. For \method{}, query-dependent evidence activation, generation of the corresponding evidence LoRA, and adaptive memory evidence decoding were included in this stage. Base Model w/ Context (oracle) re-encoded the full context history for every query, whereas parameterized methods provided only the query to the target language model during generation.

We synchronized CUDA immediately before starting and after completing each measured phase to include all asynchronous GPU operations in the elapsed time. We did not discard separate warm-up instances because the official D2L protocol does not specify a warm-up exclusion. All methods used an evaluation batch size of one in the efficiency experiments, greedy decoding, and the same maximum generation length.

Tables~\ref{tab:update-efficiency} and~\ref{tab:generation-efficiency} report update and generation efficiency, respectively, on SQuAD and 2Wiki under this unified protocol. Together, the tables separate the one-time context-update cost from the cost incurred while answering queries, allowing direct comparison among full-context inference, context parameterization, context distillation, parameter editing, and \method{}.

\textbf{Update Memory} and \textbf{Generation Memory} measure the \textbf{peak additional CUDA memory allocated} during their respective phases. For a phase $s\in\{\mathrm{upd},\mathrm{gen}\}$, we compute
\begin{equation}
M_s
=
M^{\mathrm{peak}}_s-M^{\mathrm{start}}_s,
\label{eq:phase-memory}
\end{equation}
where $M^{\mathrm{start}}_s$ is the allocated CUDA memory immediately before the phase begins and $M^{\mathrm{peak}}_s$ is the maximum allocated CUDA memory observed at any point during that phase. This additional-memory definition deliberately excludes model parameters and other allocations already resident at the phase boundary, so as to isolate the incremental cost attributable to the phase itself. For methods with multiple sequential update components, we reported the largest additional peak observed among the components rather than summing their individual peaks, since the components did not coexist in memory at the same time. We took the maximum phase-level memory value across all evaluated contexts or queries to reflect the worst-case footprint. In multi-GPU experiments, measurements were collected independently on each GPU, and the maximum across workers was reported as the representative value.

Memory was measured using \texttt{torch.cuda.memory\_allocated()} and \texttt{torch.cuda.max\_memory\_allocated()}, rather than relying on reserved memory, which can overstate actual usage. We additionally sampled process-level GPU memory with \texttt{nvidia-smi} every 200\,ms in order to independently verify the end-to-end absolute peak, but used the phase-level additional allocated memory as the basis for the reported Update Memory and Generation Memory. Under this protocol, the generation memory of Base Model w/ Context (oracle) included the additional activations and KV cache induced by the full context history, whereas parameterized methods instead measured only the additional memory required to answer the query using their previously constructed parameterized states.

\section{Data Annotation Guidelines}
\label{app:annotation-guidelines}

This section describes the annotation and verification protocols used to construct \benchmark{} and the cross-task evaluation data. The protocol follows the task definition in Section~\ref{sec:benchmark}: the complete history is retained in chronological order, later information supersedes only the corresponding earlier information, and information outside the update scope remains valid. Accordingly, annotation is performed at the level of a context and all of its associated questions, rather than by independently editing isolated question--answer pairs.

\subsection{Task Semantics}
\label{app:muse-task-semantics}

For an original context $C_o$ and update $C_u$, \task{} retains the chronological history $C_f=C_o\Vert C_u$ rather than replacing or deleting earlier text. The later context determines the current value only for information within its update scope; all unrelated information in $C_o$ remains valid. Methods concatenate $C_{\mathrm{old}}$ and $C_{\mathrm{update}}$ to form $C_{\mathrm{full}}$; the hypernetwork parameterizes $C_{\mathrm{old}}$, $C_{\mathrm{full}}$, or memory units derived from $C_{\mathrm{full}}$, but does not parameterize $C_{\mathrm{update}}$ separately. The target language model receives only the query as textual input.

Each query is assigned to $\mathcal{Q}_{\mathrm{upd}}$ if its reference answer changes under the update, or to $\mathcal{Q}_{\mathrm{keep}}$ if its answer should remain unchanged. Both sets are evaluated against the currently valid state defined by the full history. Success therefore requires simultaneously adopting revised information for $\mathcal{Q}_{\mathrm{upd}}$ and preserving unaffected information for $\mathcal{Q}_{\mathrm{keep}}$; improving one set by indiscriminately overwriting the other does not satisfy the task.

\paragraph{Recursive state evolution.}
The two-context notation above describes one transition within a recursive sequence. Starting from $C^{(0)}$, update $C_u^{(t)}$ is appended to the complete history available before step $t$:
\begin{equation}
\begin{aligned}
C^{(t)}
&=C^{(t-1)}\Vert C_u^{(t)},\\
&=C^{(0)}\Vert C_u^{(1)}\Vert\cdots\Vert C_u^{(t)}.
\end{aligned}
\label{eq:recursive-context-history}
\end{equation}
Consequently, the output history of one transition is the input history of the next: at step $t+1$, $C_{\mathrm{old}}=C^{(t)}$, $C_{\mathrm{update}}=C_u^{(t+1)}$, and $C_{\mathrm{full}}=C^{(t+1)}$. No earlier text is removed from the accumulated history. If several updates concern the same fact or dependency chain, the latest applicable statement determines its current value, while earlier versions remain in the history as invalidated evidence. Information outside the union of the update scopes retains its most recent valid value. The correction notices delimit successive transitions and make this chronological precedence explicit.

\paragraph{Recursive parameterization and evaluation.}
At every step, the method parameterizes both the pre-update history and the newly accumulated history,
\begin{equation}
\begin{aligned}
\Delta\mathrm{W}_{\mathrm{old}}^{(t)}
&=H_\phi\!\left(C^{(t-1)}\right),\\
\Delta\mathrm{W}_{\mathrm{full}}^{(t)}
&=H_\phi\!\left(C^{(t)}\right).
\end{aligned}
\label{eq:recursive-context-parameterization}
\end{equation}
and constructs the step-specific global update representation from their difference as defined in Eq.~\ref{eq:global_adapter}. Query-activated memory evidence is selected from units in the current complete history $C^{(t)}$. Thus, each transition must resolve the latest update against every version retained so far, rather than treating $C_u^{(t)}$ as an isolated document. After constructing the current parameterized state, queries are answered without supplying the textual history to the target model.

The query partition and references are also transition-specific. A query belongs to $\mathcal{Q}_{\mathrm{upd}}^{(t)}$ when its correct answer changes from step $t-1$ to step $t$, and to $\mathcal{Q}_{\mathrm{keep}}^{(t)}$ when the answer remains valid across that transition. A query unaffected at one step may become update-affected later, or vice versa. In every case, the reference $y^{\star(t)}(q)$ is determined by the latest valid evidence in $C^{(t)}$. Evaluation at step $t$ therefore measures both adoption of the new state and retention of all information that should survive that particular transition.

\paragraph{Benchmark instantiation.}
Each standard \benchmark{} example instantiates one transition, using $C_o$, $C_u$, and $C_f$ as shorthand for $C^{(t-1)}$, $C_u^{(t)}$, and $C^{(t)}$. This transition-level design supports controlled comparison across methods while preserving the recursive task definition. The continual-update experiment in Appendix~\ref{app:multi-update-experiments} explicitly chains multiple transitions, repeatedly carrying $C^{(t)}$ forward as the pre-update history and evaluating the resulting state after increasing numbers of accumulated updates.

\subsection{\benchmark{}}
\label{app:muse-annotation-guidelines}

\paragraph{Context and version structure.}
For every instance, the original context $C_{\mathrm{old}}$ is placed first and \textbf{preserved verbatim}, including sentence order, punctuation, spacing, and any source-specific formatting. We then append exactly one fixed notice, ``[Correction \& Update Notice]: The following account supersedes the corresponding details above.'', followed by the update context $C_{\mathrm{update}}$; adjacent components are separated by a single line break. This produces $C_{\mathrm{full}}=C_{\mathrm{old}}\Vert C_{\mathrm{update}}$ while making the temporal precedence relation explicit. The notice applies only to overlapping details: facts not addressed by the update remain valid and continue to support $\mathcal{Q}_{\mathrm{keep}}$.

\paragraph{Selecting the update scope.}
We first locate all evidence units supporting a candidate answer and record the complete support chain, including intermediate facts needed for multi-hop or causal inference. We select a target whose value can be changed without altering the topic, task type, or core discourse structure. The new value must constitute a genuine factual change rather than a difference in capitalization, punctuation, spelling, inflection, alias choice, numeric formatting, or unit expression. It must also preserve the answer type requested by the query (e.g., person, location, date, scalar, list, or yes/no). We avoid updates that require broad unrelated changes or create an implausible document merely to force a different answer.

\paragraph{Structure-preserving construction.}
The update context is a natural, self-consistent account rather than a short patch tailored to the selected query. It retains the source's topic, discourse order, major entities, information density, and reasoning form wherever these are not affected by the update. Paragraph, passage, section, table, and list organization is preserved when it carries semantic information. An update may be shorter because redundant wording is removed, but it may not omit major entities, events, conditions, or reasoning links simply because they are not mentioned in the target query. Conversely, unrelated material is not added to imitate the length of the source. These constraints prevent answer-bearing evidence from becoming identifiable through anomalous position, detail, or brevity.

\paragraph{Dependency closure.}
After changing the target fact, we compute its \textbf{dependency closure within the document} and synchronously revise every affected statement. This includes repeated or paraphrased mentions; aliases, abbreviations, and coreference; forward and inverse relations; intermediate multi-hop nodes; temporal order, age, duration, and date relations; quantities, totals, percentages, rankings, and unit conversions; causal consequences and conditional outcomes; and summaries, captions, tables, discussions, or conclusions derived from the target fact. All mechanically checkable relations are recalculated. No statement after the notice may preserve a direct or indirect path that makes the old answer currently valid, and the revised statements may not introduce a second plausible answer.

\paragraph{Affected and unaffected questions.}
Consistent with Section~\ref{sec:benchmark}, every question is assigned to either $\mathcal{Q}_{\mathrm{upd}}$ or $\mathcal{Q}_{\mathrm{keep}}$. For $q\in\mathcal{Q}_{\mathrm{upd}}$, the reference answer must differ semantically from the original answer, and the update context must explicitly state the complete evidence needed to obtain the new answer under the supersession rule. For $q\in\mathcal{Q}_{\mathrm{keep}}$, the original answer is retained unchanged, and neither the target update nor any dependent revision may alter its supporting evidence or otherwise render it ambiguous. Thus, $C_{\mathrm{full}}$ must simultaneously support all current answers in both sets; a document is rejected outright if fixing an affected query happens to invalidate an intended unaffected query.

\paragraph{Question and answer preservation.}
\textbf{Questions are kept verbatim} whenever they remain well formed and coherent under the new state. A question is changed only if it explicitly contains a replaced entity, value, relation, condition, or false premise, and then only the smallest coupled span necessary is revised; its intent, difficulty, answer type, and position within the set are all preserved. Duplicate questions and their relative ordering are likewise retained without modification. Answers remain concise and consistently follow the source dataset's granularity and data type. Every entity, number, qualifier, and list member appearing in an answer must be supported by the currently valid evidence, with correct event association, temporal scope, geographic level, unit, and set boundary. Mere surface occurrence of an answer string is insufficient unless the surrounding context clearly establishes the queried relation.

\paragraph{Naturalness and leakage prevention.}
Apart from the fixed notice, $C_{\mathrm{update}}$ contains only ordinary declarative prose in the style of the source. It cannot reproduce or closely paraphrase the query, introduce question--answer formatting, list responses in query order, state ``the answer is,'' refer to prompts or annotation, or describe how an earlier answer was changed. Supporting facts are inserted at their natural discourse locations. For inference-oriented examples, the update provides the necessary premises but does not append a query-specific conclusion that performs the intended reasoning for the model. The update must be understandable without external knowledge, while unrelated valid information may still be inherited chronologically from $C_{\mathrm{old}}$.

\paragraph{Dataset-specific constraints.}
For \textbf{SQuAD}, we preserve the main narrative, entity roles, and event organization, and verify identity, date, location, quantity, causality, condition, and enumeration relations separately. For \textbf{ROPES}, we preferentially preserve the background scientific or commonsense principle and modify the concrete scenario's conditions; the principle itself is changed only when no coherent scenario-level update can change the target answer. The resulting premises must still require the same type of reasoning rather than directly stating the comparison outcome. For \textbf{2WikiMultihopQA}, all source passages and their order are retained, every hop connecting the question entity to the new answer is explicit, and reciprocal family or relational statements are updated together. For \textbf{MultiFieldQA-en}, the long document's organization and topical coverage are maintained, with repeated evidence checked throughout the document. For \textbf{QASPER}, the paper-like structure is preserved and modified facts are propagated across the abstract, method, experiments, tables or captions, results, discussion, conclusion, glossary, and appendix when present; numerical totals, subsets, percentages, and method--metric--conclusion relations are recomputed for consistency.

\paragraph{Quality verification and rejection criteria.}
Each candidate first undergoes deterministic checks for parseability, schema and field types, record count, identifier preservation, exact retention of $C_{\mathrm{old}}$, notice uniqueness and boundary formatting, question--answer alignment, and preservation of duplicate items. Lexical checks flag copied questions, annotation meta-language, unchanged answers in $\mathcal{Q}_{\mathrm{upd}}$, and residual old answers including aliases, abbreviations, Unicode variants, equivalent numbers, and converted units. These checks serve only as filters: semantic verification must additionally confirm the subject--relation--object match, completeness of lists and multi-hop paths, causal and temporal attribution, dependency closure, uniqueness of the current answer, and preservation of every $\mathcal{Q}_{\mathrm{keep}}$ answer. As stated in Section~\ref{sec:benchmark-dataset}, \llmname{GPT-5.6-Sol} performs 26 rounds of verification over candidate samples. A sample is returned for revision or discarded if any affected answer lacks complete support, any obsolete evidence path remains valid after the update, any dependent fact is inconsistent, any unaffected answer changes, or the update exhibits leakage, meta-narration, structural truncation, or internal contradiction. \textbf{Only samples passing all checks are retained.}

\subsection{Human Evaluation of Main Results}
\label{app:human-evaluation}

To assess the automatic metrics used in the main results, we randomly sampled 50 contexts comprising 1,037 question--answer pairs and manually evaluated the corresponding model responses. On this subset, ROUGE-L Recall, LLM-as-a-Judge, and human evaluation scored 78.17, 72.86, and 73.22, respectively, on a 0--100 scale. The automatic scores were close to the aggregate human assessment, supporting their use as evaluation indicators; LLM-as-a-Judge was more closely aligned, differing by only 0.36 points, compared with 4.95 points for ROUGE-L Recall.

\subsection{Cross-Task Data Construction}
\label{app:cross-task-annotation-guidelines}

We apply the same chronological, structure-preserving principle to GSM8K and CRUXEval without treating them as additional \benchmark{} subsets. For \textbf{GSM8K}, the final question is split into a declarative problem context and a complete query; conditional clauses and output instructions belonging to the query remain there verbatim. The fixed notice states that the revised problem statement supersedes overlapping details. We preserve the problem type and reasoning topology, change one or more numerical conditions, and recompute every dependent intermediate quantity, total, unit, and final answer. The revised problem must remain solvable, unambiguous, and comparable in difficulty, and neither the old nor revised query is duplicated inside the accumulated context. For \textbf{CRUXEval}, the fixed notice states that the revised code supersedes overlapping details. We preserve the original prediction task and program format while making a substantive, deterministic change to the program state or computation. The revised code must be syntactically valid and executable, and the reference output is obtained from the revised program and input rather than from textual resemblance or the obsolete execution trace. For both tasks, $C_{\mathrm{full}}$ is the only source of the currently valid problem or program state, and the query retains the form and output requirements of the original task.

\section{\method{} Algorithm}
\label{app:plume-algorithm}

The complete \method{} pipeline is formalized in Algorithm~\ref{alg:plume}, where the notation directly follows Section~\ref{sec:method}. The global update LoRA is constructed once for each context history and reused across queries, whereas memory evidence activation and adaptive decoding are performed for each query.

\newcommand{\plumealgcomment}[1]{{#1}}
\newcommand{\plumealgline}[2]{%
  \makebox[\dimexpr\hsize-1em\relax][l]{#1\hfill
    \mbox{{$\triangleright$ #2}}}%
}
\begin{algorithm*}[t]
\small
\SetAlCapNameFnt{\fontsize{10pt}{12pt}\selectfont}
\SetAlgoNlRelativeSize{-1}
\SetAlCapFnt{\fontsize{10pt}{12pt}\selectfont}
\SetCommentSty{plumealgcomment}
\caption{\textbf{Detailed inference workflow of \method{}.} The algorithm constructs the global update representation, activates query-relevant memory evidence, and adaptively combines their token distributions during decoding.}
\label{alg:plume}
\KwIn{$C_{\mathrm{old}}, C_{\mathrm{update}}, q$; $f_\theta,H_\phi$; $\alpha,\beta,\delta,\lambda_{\max},\tau,T_{\max}$}
\KwOut{Generated response $y=(y_1,\ldots,y_T)$}

\tcc{\globalstage{\gur{}}}
$C_{\mathrm{full}} \leftarrow C_{\mathrm{old}} \Vert C_{\mathrm{update}}$\;
\tcc{Parameterize the pre-update and full context states}
$\Delta \mathrm{W}_{\mathrm{old}} \leftarrow H_\phi(C_{\mathrm{old}})$,
$\Delta \mathrm{W}_{\mathrm{full}} \leftarrow H_\phi(C_{\mathrm{full}})$\;
\plumealgline{$\Delta \mathrm{W}_g \leftarrow \alpha\Delta \mathrm{W}_{\mathrm{full}}
 + \beta(\Delta \mathrm{W}_{\mathrm{full}}-\Delta \mathrm{W}_{\mathrm{old}})$}%
{Eq.~\ref{eq:global_adapter}}\;

\tcc{\evidencestage{\qame{}}}
$\{m_i\}_{i=1}^{n} \leftarrow \operatorname{Segment}(C_{\mathrm{full}})$\;
$s_i \leftarrow s_{\mathrm{lex}}(q,m_i)$ for each $i\in\{1,\ldots,n\}$\;
$s_{\max} \leftarrow \max_i s_i$\;
$\mathcal{I}_{\delta} \leftarrow \{i:s_{\max}-s_i\leq\delta\}$\;
\plumealgline{$i^\star \leftarrow \max \mathcal{I}_{\delta}$}%
{latest unit within the recency margin}\;
$m^\star \leftarrow m_{i^\star}$\;
\tcc{The hypernetwork receives only the activated evidence}
\plumealgline{$\Delta \mathrm{W}_e(q) \leftarrow H_\phi(m^\star)$}%
{Eq.~\ref{eq:evidence_adapter}}\;

\tcc{\decodestage{\amd{}}}
$y_{<1} \leftarrow \varnothing$\;
\For{$t \leftarrow 1$ \KwTo $T_{\max}$}{
    \tcc{Compute the global and evidence predictions}
    \plumealgline{$p_{g,t} \leftarrow
    p_{\theta\oplus\Delta \mathrm{W}_g}
    (\cdot\mid q,y_{<t})$}%
    {Eq.~\ref{eq:global-distribution}}\;
    \plumealgline{$p_{e,t} \leftarrow
    p_{\theta\oplus\Delta \mathrm{W}_e(q)}
    (\cdot\mid q,y_{<t})$}%
    {Eq.~\ref{eq:evidence-distribution}}\;
    \plumealgline{$d_t \leftarrow
    D_{\mathrm{JS}}(p_{e,t}\parallel p_{g,t})$}%
    {Eq.~\ref{eq:js-divergence}}\;
    \plumealgline{$\lambda_t \leftarrow
    \lambda_{\max}\,d_t/(d_t+\tau)$}%
    {Eq.~\ref{eq:adaptive_gate}}\;
    \tcc{Fuse token scores with the adaptive evidence weight}
    \ForEach{candidate token $v$}{
        \plumealgline{$S_t(v) \leftarrow
        \log p_{g,t}(v)
        +\lambda_t\log p_{e,t}(v)$}%
        {Eq.~\ref{eq:fusion-score}}\;
    }
    $y_t \leftarrow \arg\max_v S_t(v)$\;
    $y_{<t+1} \leftarrow y_{<t}\Vert y_t$\;
    \If{$y_t$ is an end-of-sequence token}{
        \textbf{break}\;
    }
}
\KwRet{$y$}\;
\end{algorithm*}

\section{Discussion}
\label{app:discussion}

The experiments consistently support \rpsStrongTerm{a global-first interpretation} of continual context parameterization. The global update representation supplies a stable state of the complete context history, while activated memory evidence provides a targeted correction only when the query exposes predictive disagreement. The sensitivity results reinforce this interpretation: performance peaks at an intermediate global update coefficient and a bounded local-evidence weight, whereas excessive local weighting reduces both answer recovery and preservation of unaffected information. Query dependence therefore determines where correction is needed without making the answer depend exclusively on a retrieved local fragment.

The candidate-rank analysis also clarifies why this combination is effective. Much of the valid answer remains within the model's high-probability candidates even when it is not ranked first, suggesting that continual updates often create competition among retained signals rather than simply erasing the current answer. Global update amplification and adaptive evidence decoding act on complementary parts of this problem: the former improves the representation of the current context state, and the latter selectively resolves residual query-level competition.

\section{Future Work}
\label{app:future-work}

Our future work will proceed in two main directions. On one hand, we will investigate \textbf{more faithful and capacity-aware context parameterization methods} beyond D2L, with particular attention to reducing information loss in long contexts and across multiple successive updates. On the other hand, we will explore how to combine \method{}'s training-free mechanisms with \textbf{update-aware training objectives}, for example by using global parameter differences, activated memory evidence, and predictive disagreement as supervision for hypernetwork training and evidence routing. We believe that combining efficient inference-time adaptation with more reliable context encoding will be important for building scalable and continually updatable parametric memory.

\section{Prompts}
\label{app:prompts}

This section provides the complete rubric prompts used for the LLM-as-a-Judge evaluation described in Appendix~\ref{app:evaluation-rubric}.

\paragraph{D2 Semantic Equivalence.}
\label{app:semantic-equivalence-rubric}

\begin{figure*}[t]
    \centering
    \includegraphics[width=\textwidth]{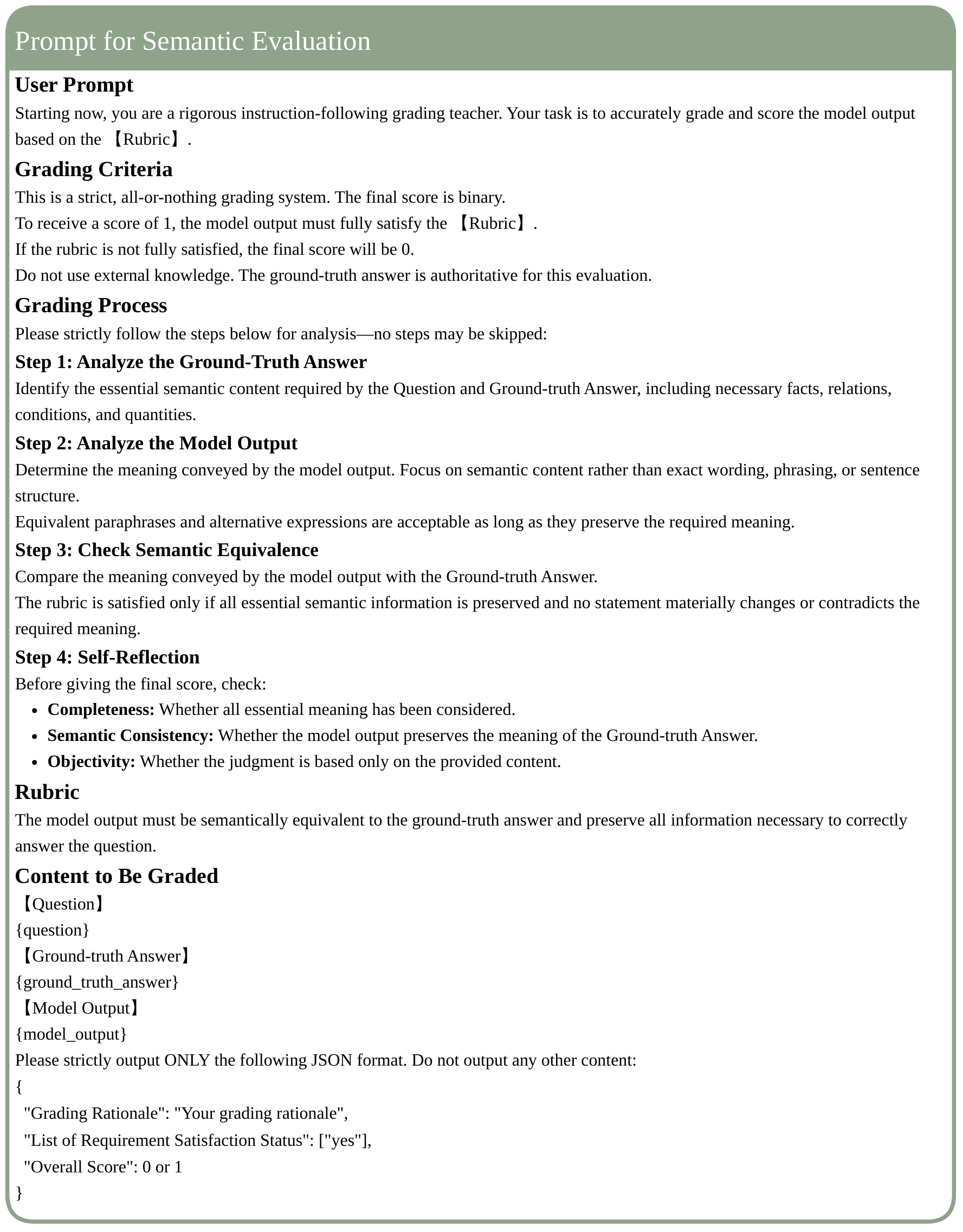}
    \caption{\textbf{Prompt used for D2 Semantic Equivalence evaluation.} The rubric determines whether a model output expresses the same answer as the current reference while preserving all required information.}
    \label{fig:semantic-evaluation-prompt}
\end{figure*}

D2 evaluates whether the model output expresses the same answer as the current reference answer and preserves all information required to answer the query correctly. The complete prompt is shown in Figure~\ref{fig:semantic-evaluation-prompt}.

\paragraph{D3 Final-Answer Agreement.}
\label{app:final-answer-rubric}

\begin{figure*}[t]
    \centering
    \includegraphics[width=\textwidth]{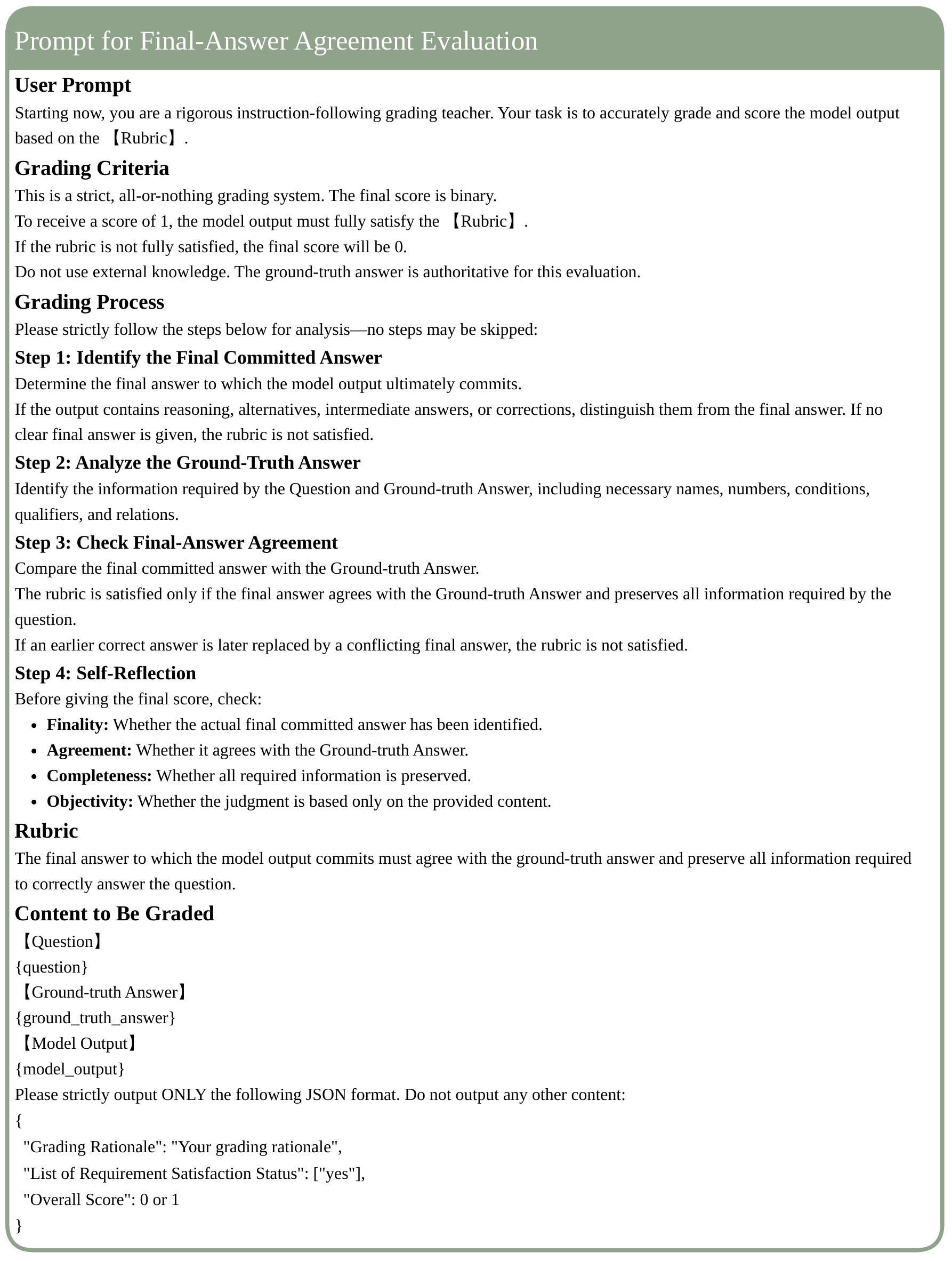}
    \caption{\textbf{Prompt used for D3 Final-Answer Agreement evaluation.} The rubric determines whether the model's committed final answer agrees with the current reference, including required qualifiers and relations.}
    \label{fig:final-answer-evaluation-prompt}
\end{figure*}

D3 evaluates whether the final answer to which the model commits agrees with the current reference answer, including all qualifiers and relations required by the query. The complete prompt is shown in Figure~\ref{fig:final-answer-evaluation-prompt}.

\end{document}